# Command A: An Enterprise-Ready Large Language Model

**Cohere**[1]

**Abstract**

In this report we describe the development of Command A, a powerful large language model purpose-built to excel at real-world enterprise use cases. Command A is an agent-optimised and multilingual-capable model, with support for 23 languages of global business, and a novel hybrid architecture balancing efficiency with top of the range performance. It offers best-in-class Retrieval Augmented Generation (RAG) capabilities with grounding and tool use to automate sophisticated business processes. These abilities are achieved through a decentralised training approach, including self-refinement algorithms and model merging techniques. We also include results for Command R7B which shares capability and architectural similarities to Command A. Weights for both models have been released for research purposes. This technical report details our original training pipeline and presents an extensive evaluation of our models across a suite of enterprise-relevant tasks and public benchmarks, demonstrating excellent performance and efficiency.

## 1 Introduction

Large Language Models (LLMs) are Artificial Intelligence (AI) models designed to understand and generate human-like text conditioned on the input they receive. Recent advancements have led to remarkable breakthroughs in their ability to comprehend and produce human language with unparalleled accuracy and fluency. This progress has been instrumental in their widespread adoption across various real-world and enterprise environments, where they significantly boost operational efficiency and deepen understanding.

This technical report describes the development of Command A and Command R7B, two LLMs designed to excel in real-world enterprise settings. Both the 111B parameter Command A and Command R7B perform best-in-class across a suite of established benchmarks for their respective model sizes. We also highlight key innovations and technical contributions including data and architectural optimisations, self-refinement algorithms, and a model merging-based approach optimised to bring out expert-level performance across capabilities within a single set of model weights, providing fast and efficient performance.

Command A is tailored for excellent performance in enterprise-relevant settings such as Retrieval Augmented Generation (RAG), where models can interact with, understand, and process information distributed across a wide range of documents. As part of this focus, our models also excel in the multilingual setting, supporting 23 key languages of global business: English, French, Spanish, Italian, German, Portuguese, Japanese, Korean, Arabic, Chinese, Russian, Polish, Turkish, Vietnamese, Dutch, Czech, Indonesian, Ukrainian, Romanian, Greek, Hindi, Hebrew, and Persian.

Along with its impressive overall performance, achieving best-in-class results for any model in its size and efficiency range on common benchmarks such as MATH, Command A outperforms across an extensive suite of human evaluation tasks as shown in Figure 1. Furthermore, Command A achieves strong results on enterprise-relevant agentic benchmarks such as Taubench, as shown in Table 1.

Command A focuses on delivering competitive performance as efficiently as possible. With a serving footprint of just two A100s or H100s, Command A requires considerably less computational overhead than comparable models. This is of particular importance for privacy-preserving enterprise settings and on-premises deployments. Command A can deliver tokens at a rate of up to 156 tokens/sec which is 1.75x higher than GPT-4o and 2.4x higher than DeepSeek V3.

[1] Please cite this technical report as "Cohere (2025)". A full author list can be found at the end of this document.

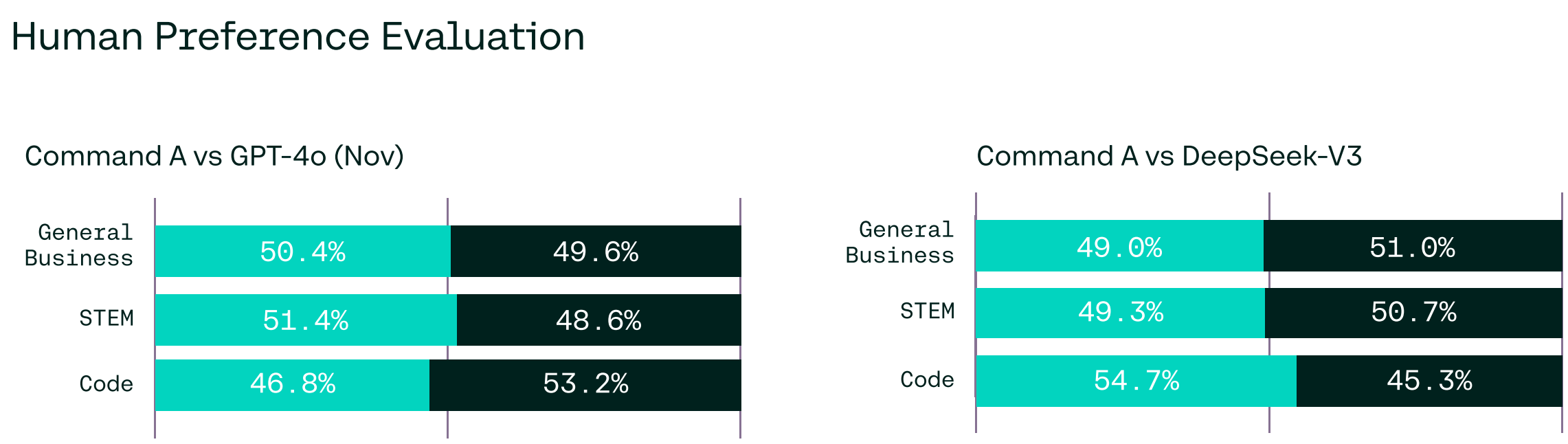

Figure 1: Head-to-head human evaluation win rates. All examples are blind-annotated by specially trained human annotators, assessing enterprise-focused accuracy, instruction following, and style.

| Capability | Benchmark | Command A | DeepSeek V3 | GPT-4o | Llama 3.3 70B | Command R7B | Llama 3.1 8B | Ministral 8B |
|---|---|---|---|---|---|---|---|---|
| **Academic** | MMLU | 85.5 | 88.5 | 85.7 | 86.0 | 65.2 | 71.1 | 71.1 |
| | MATH | 80.0 | 70.2 | 68.5 | 77.0 | 59.1 | 51.9 | 54.5 |
| | IFEval | 90.9 | 86.1 | 83.8 | 92.1 | 77.9 | 78.6 | 59.0 |
| | GPQA | 50.8 | 59.1 | 53.6 | 50.5 | 26.3 | 23.4 | 23.4 |
| **Agents** | Taubench | 51.7 | 39.1 | 51.2 | 21.0 | | | |
| | BFCL | 63.8 | 58.6 | 72.1 | 51.4 | 52.2 | 50.9 | 51.8 |
| **Code** | MBPP+ | 86.2 | 89.9 | 86.5 | 84.4 | 72.0 | 72.8 | 61.1 |
| | Bird-SQL | 59.5 | 53.1 | 50.5 | 58.0 | 42.2 | 41.9 | 33.2 |
| | RepoQA | 92.6 | 92.2 | 91.2 | 85.6 | 69.6 | 73.6 | 62.0 |
| **Multilingual** | NTREX | 68.8 | 69.8 | 71.0 | 62.5 | 48.1 | 49.2 | 36.8 |

Table 1: Command A and Command R7B results on key academic, agentic, code and multilingual benchmarks, in comparison to relevant external models.

We also release model weights to the research community to facilitate community-based exploration under a CC-BY-NC License (Non-Commercial) with an acceptable use addendum. The model checkpoints are available on the HuggingFace model hub.

# 2 Pre-training

## 2.1 Overview

Pre-training language models involves training a model on trillions of tokens of unlabelled text data to learn general language patterns, syntax, and semantics, enabling it to generate contextually relevant responses. This foundational step leverages self-supervised learning techniques, such as next-token prediction, to build a versatile representation of language that can subsequently be fine-tuned for specific downstream tasks. Pre-training is computationally intensive but essential for achieving state-of-the-art performance across diverse natural language processing applications.

## 2.2 Data

Command A models are trained on multilingual data (also see Section 3.3.3.1) from various sources including publicly available text and code data from the web, a collection of synthetic datasets generated internally, instruction-tuning datasets obtained from human annotators, and high quality data sourced from specialised data vendors. We optimise the web text data by enhancing the ratio of educational samples that are relatively sparse on the internet, and down-sampling low-quality samples identified by Machine Learning (ML)-based quality filters after careful de-duplication and heuristic filtering for safety and quality. The final data mixture is determined by running a series of ablations using smaller models.

## 2.3 Model Architecture

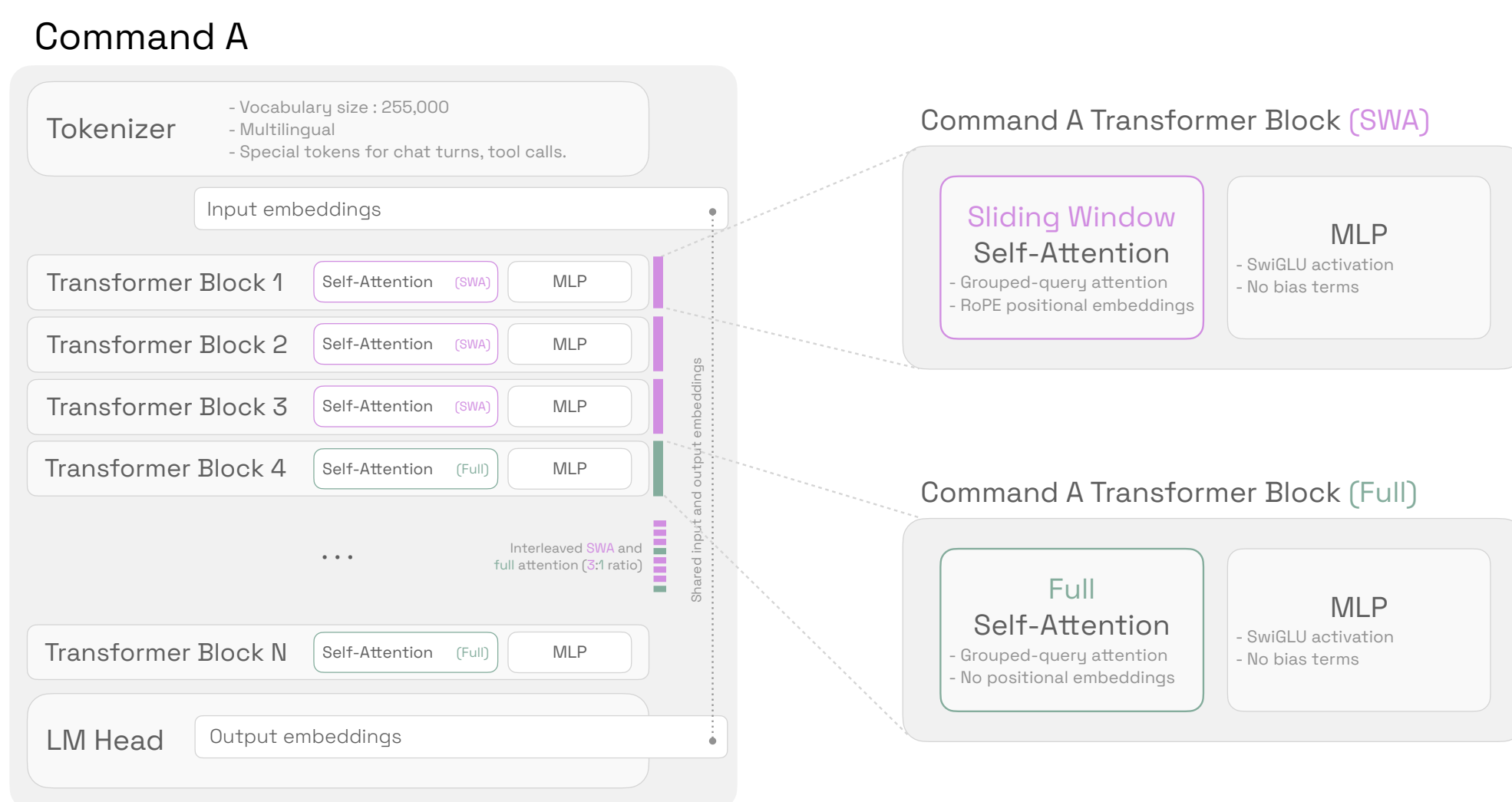

Figure 2: Schematic of the Command A model architecture.

We use a decoder-only Transformer architecture (Vaswani et al., 2017) as depicted in Figure 2. We highlight a few key architectural decisions below:

- **SwiGLU.** The SwiGLU activation (Shazeer, 2020) demonstrates performance improvements over other activation functions.
- **Interleaved attention layers.** We use interleaved layers of sliding window attention and full attention in 3:1 ratio. Each sliding window layer uses Rotary Positional Embeddings (RoPE) (Su et al., 2021) and every full attention layer uses No Positional Embeddings (NoPE) (Kazemnejad et al., 2023). Further details can be found in Yang et al. (2025).
- **GQA.** We use grouped-query attention (GQA; (Ainslie et al., 2023)) to increase serving throughput. We use document masking to ensure that each individual sequence in a batch can only attend to itself.
- **Parallel transformer block.** This shows equivalent performance but significant improvement in throughput compared to the vanilla transformer block.
- **No bias.** Similar to PaLM (Chowdhery et al., 2023), we do not employ bias terms, which improves training stability at larger scales.
- **Input and output embeddings.** We share the input and output embedding matrices, which provides a large reduction in memory requirements due to our large vocabulary size. We do not observe any performance degradation across ablations.

## 2.4 Pre-training Recipe

We perform most of our hyperparameter optimisation at considerably smaller scales than those representing our final family of models. We use $\mu$P and $\mu$Transfer (Yang et al., 2021) to tune hyper-parameters on smaller models and zero-shot transfer them to our larger models. Sweeps are performed for each model size as they assume a fixed number of layers.

### 2.4.1 Distributed Training

We train all our models on our NVIDIA H100 GPU cluster using our internal JAX-based (Frostig et al., 2018) distributed training framework. Our framework leverages JAX's GSPMD (Xu et al., 2021) as the backbone to implement complex sharding strategies. We split the available GPUs into a mesh with four axes for each of the following sharding schemes:

- **Data Parallel (DP)** axis shards the activations along the batch dimension, which behaves as standard data parallel training when all GPUs are allocated to it.
- **Fully Sharded Data Parallel (FSDP)** axis shards both the activations along the batch dimension and model states along a specified dimension. The model states are replicated across the data parallel axis to contain the communication costs to the FSDP submesh.
- **Sequence Parallel (SP).** Given the restrictions on critical batch-size of LLMs, scaling the number of GPUs in pure FSDP/DP scenarios is infeasible. We thus use sequence parallelism (Li et al., 2023b) to shard activations along the sequence dimension. The activations after the QKV projection are sharded along the heads dimension to remove communication costs during the attention computation. The attention outputs are sharded on the outer dimension, and the weight matrix of the final attention output transformation is sharded along the contracting dimension as in Megatron-style (Shoeybi et al., 2019) sharding. This allows us to operate using a single all-gather and a single reduce-scatter for the activations, while only gathering QKV and attention outputs along the FSDP axis. At the feed forward block, the FFN transformation is independent along the sequence axis, therefore there is no need for any activation communication. Moreover, since we use parallel attention and a FFN block setup, we completely overlap the computation of the FFN expansion layer and the all-gather of the attention activations. The reduce-scatter after the attention block is further overlapped with the execution of the FFN reduction layer. Since all other major operations such as layer norm, input and output embedding layers are independent along the sequence axis, they incur no communications along the activations.
- **Tensor Parallel (TP)** axis for a pure Megatron-style sharding, where two complementary matrix multiplications are sharded such that the activations are all-gathered before the first matrix multiplication (where the weight is sharded on the outer axis, resulting in the activations being sharded on the outer axis as well), and one all-reduce after the second matrix multiplication (to sum the partial outputs). Pure TP is desirable when moving activations between devices as it is much cheaper compared to moving weights, a layout class sometimes referred to as weight-stationary (Pope et al., 2023). We use pure TP for fast decoding and in low batch-size scenarios.

Our models are trained with varying combinations of the parallelism strategies mentioned above. During pre-training, since we are in the high batch-size and throughput regime, we opt for a combination of DP, FSDP and SP to minimise activation communication. Furthermore, we can unroll the model's forward loop to overlap the communication of the weights of the next layer with the execution of the current layer.

We leverage Hopper-specific features such as FP8 tensor cores (Micikevicius et al., 2022) to further improve throughput. While many works have reported instability while training with FP8 precision for long training horizons (Fishman et al., 2025), we observe no such instability. In fact, we observe minimal run interventions due to loss spikes and optimisation instability. We keep our main weights and optimiser states in FP32 precision, and cast the model weights to BF16 or FP8 prior to the computation. We keep sensitive operations such as exponentials, softmaxes, layer norms, and output embeddings in FP32 precision, and run the attention computation in BF16 precision. While we do not observe any training instabilities with FP8 matmuls, we notice that there is a small but non-trivial degradation in downstream performance if the entire training run is in FP8. To mitigate this effect, we first perform a number of steps in BF16 precision, which brings performance back to the full BF16 trained model's performance range.

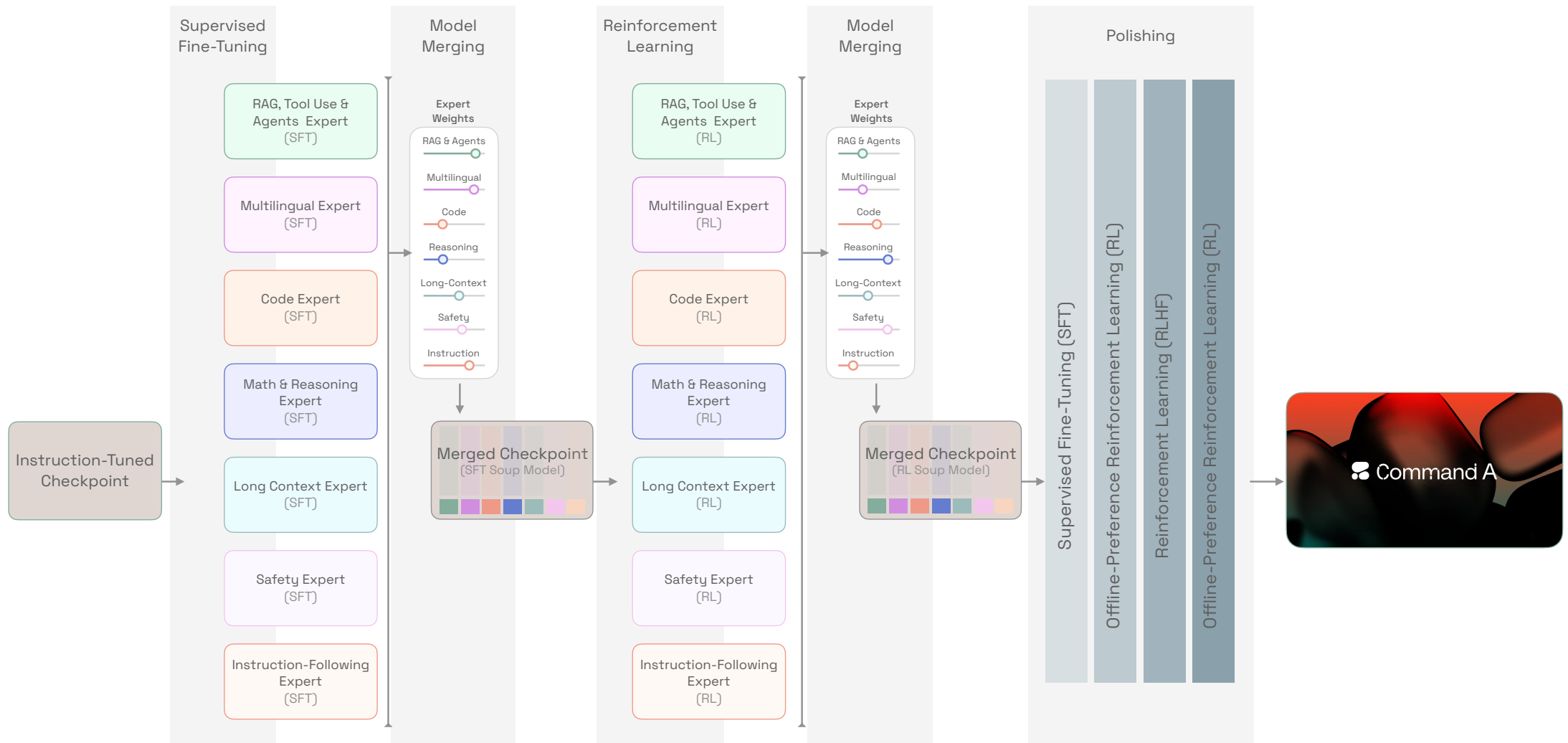

Figure 3: Command A goes through multiple post-training phases including two weighted model merging steps, and a model polishing phase.

## 2.5 Cooldown

We linearly anneal the learning rate from $2.5 \times 10^{-4}$ to $1 \times 10^{-6}$ for 50,000 steps in BF16 precision with purposely curated high quality datasets. The context length is initially maintained at 8k tokens for the first 30,000 steps, then extended to 32k, 128k, and 256k for 10,000, 5,000, and 5,000 steps respectively by interleaving long context pre-training data every fourth step. During the long-context stages, we adjust the overall ratio of datasets to ensure a balanced mixture across domains (Fu et al., 2024), while maintaining a sufficient proportion of long-context data.

# 3 Post-training

## 3.1 Overview

Command A is post-trained using a novel decentralised approach to maximise and control its performance over a wide spectrum of domains and capability areas. More precisely, Command A is trained by alternating centralised training stages, where a single model is fine-tuned, and decentralised training stages, where multiple expert models are trained separately to maximise domain-wise performance before merging their parameters. Although the classic post-training paradigm involves training a single model sequentially with varying data-mixtures (Dubey et al., 2024; Team et al., 2024), Command A is the first large-scale attempt to combine multiple parallel expert tracks, with parameter merging techniques playing a central role. This section details the high-level post-training recipe of Command A, illustrated in Figure 3.

We divide the global Command A post-training recipe into several sub-stages, each producing intermediary model artifacts:

- **Instruct Model:** We train an initial Instruct model with supervised learning on top of the base model to provide the core basic capabilities of the model.
- **SFT Expert Models:** We train six SFT experts on top of the Instruct checkpoint with specialised data mixtures to maximise capability-specific performance.
- **SFT Soup Model:** We merge the six model experts into a Soup model with parameter-merging methods (see Section 3.4) to produce a single SFT aggregate model.

- **RL Expert Models:** We train six RL experts on top of the SFT Soup checkpoint using RL algorithms tailored to each domain, using pairwise comparisons or verifiable rewards.
- **RL Soup Model:** We merge the six RL experts into a RL Soup model with parameter-merging methods to produce a single RL aggregate model.
- **Polished Model:** We perform a final stage on the RL Soup model to enhance human interaction performance by alternating between best-of-N methods, offline preference, and online RL algorithms.

Six expert models are created at each expert stage: Code, Safety, RAG, Math, Multilingual, and a General Long-Context expert. This approach allows us to adapt each expert's training procedure, tailoring it to the specific capability or domain of interest. This allows fine-grained hyperparameter tuning, specialised data mixture optimisation, local optimisation (e.g., seed merging), and the capability-specific selection of the most appropriate algorithms. This becomes even more crucial during the RL stage, as different domains demand distinct RL techniques — for example, verifiable rewards for Math and Code, or preference pairs for Safety and Multilingual. Our late-merging procedure allows us to re-balance Soup model performance *a posteriori* without requiring additional training (§3.4). From an organisational perspective, merging allows contributors to collaborate closely in parallel, fostering a unique model development synergy. Overall, this decentralised training procedure maximises individual expert performance while controlling the final global model capacity, allowing us to optimise both model performance and efficiency.

Finally, the model undergoes a polishing phase to improve its writing style. First, we apply a best-of-N supervised training stage to the RL Soup model. Then, we alternate between offline preference and online RL optimisation in a ping-pong approach, iterating as required until we observe a human preference performance plateau, to obtain the final Command A model.

In the following sections, we introduce the methods and algorithms that we use at various stages of the post-training process. We detail individual expert recipe considerations and provide further technical details on the merging techniques applied. Finally, we discuss key features of the model polishing phase.

## 3.2 Methods

### 3.2.1 Supervised Fine-Tuning

In all cases, the first stage of our post-training pipeline involves finetuning the pretrained model to follow instructions and operate in a conversational setting. We structure Supervised Fine-Tuning (SFT) (Wei et al., 2021; Sanh et al., 2021) datapoints as prompts and completions. Prompts are model input sequences that may contain information such as preambles or system prompts defining expected model behaviour, tool specifications, conversational history, special tokens (e.g., `<|START_OF_TURN_TOKEN|>` or `<|SYSTEM_TOKEN|>`), and queries or instructions. Completions refer to the sequence of tokens that the model is trained to generate conditioned on a given prompt. We train the model using a cross-entropy loss with the loss corresponding to prompt tokens masked out. Depending on the specific setting, we may choose to regularise training either by including some small proportion of pretraining data, or a parameter-based $L_2$ penalty to the pretrained model. We optimise using the AdamW algorithm (Loshchilov & Hutter, 2017) with decoupled weight decay.

### 3.2.2 Reinforcement Learning

Depending on the stage and task, we perform direct alignment using preference training (Rafailov et al., 2024; Azar et al., 2024) or we optimise for a reward signal through reinforcement learning (RL) (Sutton et al., 2023), either offline or online. This reward signal can be the learnt reward model, or a verifiable reward (for example, based on unit tests for code generation or on response correctness for reasoning).

#### *3.2.2.1 Preference Training with Self-refinement*

We consider preference training methods for learning offline from preference datasets: Sequence Likelihood Calibration, or SLiC (Zhao et al., 2023), Direct Preference Optimisation, or DPO (Rafailov et al., 2024), and Identity Preference Optimisation, or IPO (Azar et al., 2024). In addition to these conventional preference-training methods, our model-training pipeline incorporates our novel Self-improving Robust Preference Optimisation (SRPO) (Choi et al., 2025). This recently-developed approach represents a significant departure from traditional preference training techniques, introducing a novel mechanism for continuously enhancing model alignment and robustness. It amounts to solving the following min-max optimisation problem:

$$\min_{\pi} \max_{\pi_\dagger} \mathbb{E}_x \mathbb{E}_{y_1 \sim \pi(\cdot|x), y_2 \sim \pi_\dagger(\cdot|x,y_1)} \left[ P(y_2 \succ y_1|x) - \beta \, \mathrm{KL}(\pi_\dagger || \pi_{\mathrm{ref}}|x, y_1) + \beta \, \mathrm{KL}(\pi || \pi_{\mathrm{ref}}|x) \right].$$

This objective function aims to learn a self-improvement policy $\pi_\dagger$ that can improve generations from $\pi$, according to the preference model $P$ without deviating too much from a reference model $\pi_{\mathrm{ref}}$, and at the same time at learning a policy $\pi$ of which generations cannot be improved by $\pi_\dagger$. SRPO's novelty partly lies in its robustness: unlike classical methods, it does not depend on the sampling distribution of the preference dataset. This independence ensures greater generalisation and stability in varied deployment scenarios. Furthermore, SRPO uniquely enables iterative self-revision at inference time, a process where the model sequentially refines its output: Given an initial prompt, the model first generates an initial completion using the generative policy $\pi$, followed by multiple sequential refinements through the self-refinement policy $\pi^\dagger$, each progressively improving the quality and alignment of the final output. This iterative refinement capability is not present in conventional alignment pipelines, underscoring SRPO's innovative contribution.

##### 3.2.2.2 Optimising the Reward Model with RL

When given a reward function, be it the reward model or a verifiable reward, we consider the classic KL-regularised reinforcement learning objective, $J(\pi) = \mathbb{E}_x \mathbb{E}_{y \sim \pi(\cdot|x)}[R(x, y) - \beta \, \mathrm{KL}(\pi || \pi_{\mathrm{ref}}|x)]$. In all settings (offline or online), we optimise it using the recent Contrastive Policy Gradient approach, or CoPG (Flet-Berliac et al., 2024). For a prompt $x$ and $k > 1$ completions $y_1, \ldots, y_k$ of arbitrary origin, the corresponding loss to be minimised is

$$\ell(x, y_1, \ldots, y_k; \pi) = \frac{1}{k-1} \sum_{i>j} \left( R^\pi_\beta(x, y_i) - R^\pi_\beta(x, y_j) \right)^2 \text{ with } R^\pi_\beta(x, y) = R(x, y) - \beta \ln \frac{\pi(y|x)}{\pi_{\mathrm{ref}}(y|x)}.$$

The CoPG loss can be used in both the offline and online cases. In the online case, it can be used with a replay buffer, possibly combined with additional datasets, or in a pure on-policy fashion, in which case it is equivalent to Reinforce Leave-One Out, or RLOO (Ahmadian et al., 2024). Furthermore, Flet-Berliac et al. (2024) show that the gradient of this loss is a form of (negative) off-policy policy gradient, not relying on importance sampling, clipping, or on an additional value network. It also comes with strong theoretical guarantees, notably estimating the KL-regularised optimal policy $\pi^*$ even in the offline case, and generalising policy gradient and some preference training approaches. In the offline case, it can be used with any dataset, as long as there is more than one completion per prompt, and that we can compute the associated rewards. We mostly use CoPG offline and online on-policy.

#### 3.2.3 Reward Models

We train a Bradley-Terry reward model for use in online preference learning, evaluation, and data filtering. Similar to Gunter et al. (2024), we use a cross-entropy loss with soft labels as targets.

We find that reward models tend to suffer from high memorisation, causing catastrophic collapse in performance on a second epoch over the same data; so, we train the model in two stages. The first stage consists of approximately 4 million samples designated as "lower quality" and relabelled using an ensemble of reward models. Training is carried out for one epoch with a batch size of 1024 with a cosine learning rate schedule with a peak of $4 \times 10^{-5}$. The second stage consists of approximately 350,000 high quality samples, with labels derived from the strength of human preferences, ensembles of models (with labels inconsistent with human annotations moved to the first stage), or a constant label value of 0.999 for gold-standard data and 0.5 for gold-standard tied pairs. This stage uses a smaller batch size of 16, and a lower maximum learning rate of $3 \times 10^{-6}$. Both stages use packed data, where multiple pairs of preference data are encoded in a single training sample for efficiency, using attention masking to avoid different (non-packed) samples influencing each other. The pairs are left-padded to align their <EOS_TOKEN>, and distributed to aim for a 75% fill while keeping the number of samples per row as uniform as possible, ensuring an equal loss contribution.

Our internal reward model scores 92.7% on RewardBench (Lambert et al., 2024), and achieves an average score of 72.3% on RMB (Zhou et al., 2024).

## 3.3 Capabilities

### 3.3.1 Instruction-Following

Core instruction-following capabilities are crucial for LLMs to solve tasks across areas and domains. We therefore consider instruction-following a prerequisite for more specific model capabilities focusing on advanced topics such as code, multilingual, and reasoning. As such, in the Command A post-training recipe, we teach the model to follow instructions across a wide range of topics and domains, including but not limited to generalist instruction-following (e.g., factual knowledge requests), formatting, STEM-specific tasks (e.g., tabular reasoning, structured data manipulation), and preamble compliance. Instruction-following capabilities are acquired both via SFT and offline preference tuning.

#### *3.3.1.1 Data collection*

Our data collection approach can be divided based on the post-training method, i.e., SFT or preference tuning. To collect datasets that serve both of these, we primarily rely on synthetic prompt generation in conjunction with human annotation, and explore various sampling and filtering techniques (Bartolo et al., 2021). Specifically, we synthetically generate diverse sets of prompts covering a range of instructions tailored to individual domains (such domains are mostly enterprise-oriented) and generate two completions per prompt sampled with different temperatures. We then ask human annotators to provide discrete ratings for both completions. This process is repeated over multiple turns, resulting in a multi-turn dataset. If the two completion ratings are not tied and the better completion does not obtain the highest possible rating, we ask the annotator to improve the better completion.

**SFT samples.** SFT datasets are constructed using the human rewrites obtained from the process mentioned above, to ensure the highest completion quality.

**Preference pairs.** We construct preference pairs directly from the obtained samples by considering completions with different ratings (including the human rewrites), with ties excluded. It is worth noting that the obtained preference samples are used to train both Command A and our reward model itself.

#### *3.3.1.2 Iterative Reward-based Sample Refinement*

We further experiment with reward-based sample refinement approaches to obtain both SFT and preference pairs using the synthetically-generated prompts. Similar to Dubey et al. (2024), we use internal reward models trained on our most recent Command A checkpoints in conjunction with a collection of both human-written completions and completions generated from different checkpoints under different conditions (e.g., varying temperature values) during post-training This implies that the resulting dataset does not contain purely synthetic completions, and human-written completions are retained for inputs where the models fail to generate high-quality completions. We approach this in an iterative fashion, where we use the most recent checkpoints at a given point in time to generate completions, score those completions using our reward model, create preference pairs and SFT samples using the scores, re-train our models, and repeat.

#### *3.3.1.3 Preambles*

A specific focus of the instruction-following post-training of Command A lies in the model's ability to follow preamble (or system prompt) requirements. Preambles are designed to contain instructions that should apply to an entire conversation and potentially multiple model inputs, meaning that instead of having to repeat instructions in every prompt, they can be defined directly in the preamble. Such system instructions could specify what language the model should reply in (e.g., "Always respond in French."), the desired format of model generations (e.g., "Always use JSON.", "Do not generate Markdown."), or the exclusion of specific words and phrases (e.g., "Do not use the word 'LLM' in your response."). To train Command A to follow preamble instructions, we develop methods based on synthetic data generation to create diverse preambles that are attached to prompts flowing into the above-described pipeline. The preambles are then taken into account when creating the respective completions and preferences, i.e., preamble-augmented data is used during both SFT and preference tuning. During preamble generation, we aim to maximise instruction diversity to encourage robustness to a wide range of instructions at inference time.

#### 3.3.1.4 Model training

In the context of instruction-following we post-train Command A in sequence with SFT and preference tuning. For preference tuning, we experiment with a range of methods, including SLiC, IPO, DPO, and SRPO (for further details, see Section 3.2.2). We find that SRPO performs best across evaluation tasks and select a checkpoint trained using SRPO for the final Instruct model.

### 3.3.2 Retrieval Augmented Generation, Tool Use and Agents

Recent breakthroughs have propelled LLMs beyond simple chatbots, transforming them into versatile agents capable of navigating complex environments. At the heart of this evolution is their ability to use tools strategically: invoking APIs, analysing results, and iterating dynamically to accomplish goals. This agentic behaviour is pivotal for two key advancements. First, integrating knowledge sources outside of model parameters (e.g., via Retrieval-Augmented Generation, or RAG) mitigates hallucinations and ensures accurate information beyond the timespan of model training. Second, agentic frameworks empower models to orchestrate vast action chains, potentially executing hundreds of API calls to automate intricate workflows. Together, these capabilities expand LLMs' operational horizons, enabling them to tackle tasks once deemed beyond their reach.

#### 3.3.2.1 Agentic Tool-Use

**Empowering LLMs with Tools.** LLMs have demonstrated remarkable proficiency in leveraging external tools to enhance their capabilities (Schick et al., 2023). By generating API calls, models can execute specific actions—like performing calculations or retrieving information—to solve tasks more effectively. This process typically involves providing a set of tool definitions in the model's preamble. When faced with a task, the model selects and invokes the appropriate tools, and the results are fed back to inform its final response.

A prime example of this is Retrieval-Augmented Generation (RAG). In **RAG** (Lewis et al., 2020), the model has access to a search tool (e.g. a dense embedding index) to answer information-seeking queries. It generates search queries, retrieves relevant snippets from the selected knowledge source, and uses this context to craft a well-informed response.

**Agents.** For more intricate tasks, models may need to orchestrate multiple tools across several steps. This requires a structured approach to halt generation, extract tool calls, execute them, and reintroduce results into the model's workflow—a process repeated until the task is resolved.

We roughly follow the ReAct framework (Yao et al., 2022), a widely adopted method for guiding LLMs through dynamic problem-solving. ReAct enables models to interleave reasoning and action: they first articulate their thought process, outlining plans and tool requirements, then either execute a tool (via structured outputs like JSON) or deliver a final answer. This iterative loop enables adaptive planning, reflection, and interaction with external systems, making it ideal for complex, multi-step tasks.

#### 3.3.2.2 Data and Training

We train our model on a combination of human-annotated and synthetically generated data. We collect datapoints in multiple languages to directly supervise on, as well as datapoints with preference judgments for multiple completions. The data covers areas of code tool execution, user-uploaded documents, and general API environments. Training consists of an SFT step on agentic datasets followed by offline preference tuning using the Contrastive Policy Gradient loss (§3.2.2.2).

**Data format.** Each datapoint contains a user prompt along with a set of available tools and potentially custom model instructions that the user has provided to the model. The datapoint also contains a reasoning step, where the model reasons about which tools to use to fulfil the user request and how to fill in the input parameters of the tools. This is followed by tool calls and tool outputs, which can be concurrent or sequential. The datapoint concludes with a model response that includes citations to the tool outputs.

**Data collection.** Annotation is performed by internal annotators specifically trained in ReAct-style data. All annotated data used for SFT is reviewed multiple times by different annotators to ensure correctness and quality. For preference data, we use a majority vote of at least 3 annotators to collect preference judgments.

**Synthetic data.** We also generate synthetic training data containing whole trajectories of reasoning and tool calls. We verify the quality of the trajectories using internal LLMs-as-a-judge.

### 3.3.3 Multilingual

The ability to communicate in and understand multiple languages is a fundamental component of enterprise LLMs. Command A is designed to handle a wide array of languages, ensuring that information can be accessed and shared seamlessly across different linguistic communities. By incorporating solid multilingual capabilities in 23 languages, Command A enables businesses and individuals to reach a broader audience, fostering inclusivity and accessibility. Moreover, the multilingual aspect of Command A facilitates better understanding and collaboration among international teams, driving innovation and efficiency.

#### *3.3.3.1 Data Mixture*

We focus our data mixture on 23 languages: English, French, Spanish, Italian, German, Portuguese, Japanese, Korean, Arabic, Chinese, Russian, Polish, Turkish, Vietnamese, Dutch, Czech, Indonesian, Ukrainian, Romanian, Greek, Hindi, Hebrew, and Persian. This ensures coverage to expand state-of-the-art language modelling capabilities to approximately half of the world's population (Aryabumi et al., 2024; Üstün et al., 2024). Our multilingual data mixture spans a diverse set of domains and tasks, covering machine translation, multilingual safety, multilingual reasoning, multilingual robust controllability, multilingual RAG, multilingual agents, etc; ensuring that Command A possesses strong generalisation capabilities across languages. The datasets are collected through various means including human annotation, multilingual data arbitrage (Odumakinde et al., 2024; Dang et al., 2024), templated public datasets, or machine translation. Our data mixture is specifically tailored to handle multilingual learning through SFT and offline preference tuning.

#### *3.3.3.2 Multilingual Data Annotation*

Multilingual data annotation is performed by internal and external multilingual annotators who are expertly trained for annotation within various tasks. It can be divided into two distinct processes, i.e., regionally-relevant data annotation and complex multilingual task annotation, which cover use cases for both SFT and preference tuning. For complex tasks such as domain-specific RAG, long-context reasoning, or agentic tool-use tasks, we conduct human annotation using two different approaches: 1) LLM-generated response with human post-editing; and 2) manually annotated human data. The prior helps us scale the quantity of data, while the latter helps develop high-quality multilingual data for tackling complex tasks. We develop a customised in-house data annotation platform that can support both of these use cases. The high-quality data generated from human annotations also helps to further improve the quality of the machine-generated responses providing a positive feedback loop within the annotation process.

**Multilingual Best-of-N.** To further improve the multilingual quality of Command A, we conduct iterative synthetic data collection through multilingual best-of-N (Stiennon et al., 2020; Eisenstein et al., 2024). Using a collection of high-quality prompts, we collect responses from all expert models and score them using our internal reward models and select the best response to be used in our iterative training. This approach is very similar to multilingual arbitrage where the model is trained on responses from diverse teacher models. Using this approach, we observe from human evaluation that LLMs can produce responses that are comparable or even better than the human-written gold label provided in many multilingual datasets.

#### *3.3.3.3 Training*

The multilingual expert model is trained via both SFT and Preference Tuning (full details in Appendix B.2). We find that training several models with the same configuration (but a different random seed) and uniformly merging them gives a slight performance boost for the expert at the SFT stage, but does not help at the preference tuning stage.

#### 3.3.4 Code

Generating and understanding code is a fundamental requirement for any enterprise LLM. We invest in the code capabilities of Command A to assist the software development cycle and improve user coding experience. Command A's success in code-oriented tasks is also a precise measurement of capability in instruction-following, pragmatic inference in interpreting prompts, and procedural reasoning. Our models excel in challenging business environments, including understanding and translating legacy programs in COBOL, and using SQL to interface with relational databases.

##### *3.3.4.1 Data*

**Data Mixture.** Our data mix focuses on 8 priority programming languages (Python, Java, C++, Rust, Go, JavaScript, TypeScript, COBOL) and 5 dialects of SQL (SQLite, MySQL, PostgreSQL, Microsoft T-SQL, Oracle PL/SQL). Across these languages, we target a wide range of tasks including code generation (i.e., NL-to-code), code translation, and code optimisation. Within these tasks, we include diverse domains such as interview-style questions; repository-level queries; and enterprise-specific demands (including high-dimensional arrays, complex debugging, data processing, and visualisation).

Prompts and completions are sourced from annotation campaigns and synthetic generation pipelines. We enrich prompt-completion pairs with additional information including execution feedback, explanations, paraphrases, stack traces, database context (Chang & Fosler-Lussier, 2023), diff patch formats, and unit-testing requirements. We prioritize candidate data with positive execution-based validation to filter erroneous or unverifiable code. This includes passing gold-standard unit tests or correct and error-free execution within a database. We use a multi-language code execution sandbox to evaluate code correctness in an isolated environment similar to Guo et al. (2024) and Team et al. (2025).

During pretraining, we perform execution-based code data enrichment. We isolate self-contained functions and files and add print statements of some variables and generate synthetically valid input parameters. The resulting code is executed in a sandbox and the output is appended to the enriched source code, adding several billion pretraining tokens. There is the added benefit that a subset of code repositories can be formatted as a very long document where files are linearised following a graph traversal defined by import links.

In the RL stage, we jointly optimise for code correctness and annotated preferences between code completions. This approach enhances both the functional accuracy of generated code and reduces edit times of technically correct but suboptimal or dispreferred generations. We quantify performance using the proportion of unit tests passed in our code execution sandbox, where a reward of 1.0 indicates 100% test success and 0.2 represents 20% test success. When no valid code block is detected, we assign a reward of $-1.0$ to explicitly penalize non-code output. We use synthetic unit-test generation ensuring all code completions have a minimum of 4 tests per sample. Our synthetic test generation pipeline is similar to Zeng et al. (2025) with more robust `unittest` tests over `assert` statements. The preferred SQL completions are canonicalised using static query analysis. Beyond verifiable metrics, we incorporate DPO-style preference pairs (Rafailov et al., 2024) to optimise for code style conventions, documentation structure, and formatting consistency.

**Synthetic Data.** We experiment with synthetic data pipelines for post-training data enrichment. As a result, a high proportion of our data are verified synthetic examples in many coding languages. For synthetic generation sources, we exploit our highest performing models for code, and generalist models for explanations and reasoning. We experiment with both novel data synthesis and conditional synthetic data augmentation.

Our novel data synthesis efforts include generating examples taking inspiration from concepts, similar to StarCoder (Wei et al., 2024), and sampling pretraining programming queries. We explore pipelines where our synthesis is Python-only followed by translating code and unit tests into additional languages, or direct generation into any programming language. While the former is useful for generating parallel corpora and targeting Python benchmarks, the latter pipeline proves valuable for problems using absent or uncommon features of Python (e.g., multithreading or memory management for C++). We use our execution sandbox to verify all synthetic completions — ensuring that any synthetic example teaches a novel skill via verified code. This approach to data synthesis only improves performance for small models (i.e., data-based distillation from larger models). Novel synthesis methods yield negligible improvement for larger models, instead requiring human annotation and synthetic data augmentation to advance our most capable coding experts.

We rely on synthetic data augmentation to diversify the style, syntax, and complexity of our code data. Our data augmentation pipeline includes prompt paraphrasing, injecting stricter requirements into prompts for more precise instruction following, and complexifying prompts similar to Magicoder (Wei et al., 2023). In verifiable scenarios, we use execution feedback to build data for code repair or translation, where iterative feedback provides guidance until the repaired code passes all tests. In a similar scenario for SQL, a repaired or translated SQL is adequate when it returns an equivalent answer from a target language database. This offline pipeline can generate prompt-completion pairs, but we also cast this iterative process into multi-turn data to simulate conversational code repair.

We also use synthetic augmentation to improve non-verifiable aspects of our data. This includes code explanations, markdown style formatting, technical precision, and global completion structure. We use reward modelling and majority voting to score these non-verifiable code completions. We also elicit feedback from human annotators to guide our data synthesis pipeline towards developer preferences for code style, structure, and explanations. This regularises against overfitting to the preferred style of any LLM judge, and our generations' target style and structural features are actually preferred by human raters.

##### *3.3.4.2 Training*

The code expert is trained in three stages (hyperparameters and full details in Appendix B.3):

**Stage 1** is large-scale supervised learning, with the code data mixture described above. This stage includes data for all relevant tasks to optimise a high level of coding capability. To mitigate variance in initialisation, learning trajectory, and performance on small evaluation datasets we use linear merging over the top $k$ seeds (Izmailov et al., 2018; Team et al., 2024; Yadav et al., 2024; Aakanksha et al., 2024; Khalifa et al., 2025) where $k$ is typically 2 or 3. We observe that this merged model is a strictly superior initialisation for continued training with additional fine-tuning or RL.

**Stage 2** is supervised learning on only high-quality data. From the first stage fine-tuning, we further strengthen our code expert with additional fine-tuning on our "highest-quality" code generation datasets. We define "high-quality" as verifiable human or synthetic data from our best experts, or data rated highly by internal reward models. As before, we train multiple candidate models and merge across random seeds to produce the final SFT code expert. This secondary fine-tuning stage increased our key benchmark performance with negligible regression in tasks only present in stage 1 training (e.g., SQL query optimisation).

**Stage 3** is RL over scored or preference-pair data. We train the expert with the offline Contrastive Policy Gradient algorithm (§3.2.2), to train with execution feedback and DPO-style preference pairs as described above. To ensure stable RL, we introduce three regularisation schemas. First, we repeat the Stage 2 high-quality supervised learning and merging process on any non-code expert model (e.g., a merge of multiple experts). CoPG on a merged checkpoint was strictly more stable and yielded better results than RL on top of an individual SFT/merge. Second, we introduce a hybrid cross-entropy loss function on top of CoPG to sample steps of typical supervised learning from the same Stage 2 data mix. Third, we use WARP-style merging (Ramé et al., 2024) to combine the final model trained with RL to the parent checkpoint. This hybrid approach ensures stable reinforcement learning optimisation to improve both our code experts for user preference, and improving our performance on intrinsic code generation capabilities.

#### 3.3.5 Math and Reasoning

Sophisticated reasoning abilities are a necessary competency area for generalisation in LLMs (Guo et al., 2025; Team et al., 2025; Toshniwal et al., 2024). We focus primarily on core mathematical reasoning as it is both intrinsically useful (e.g., in financial use cases) and yields out-of-distribution improvements in other knowledge-intensive tasks such as coding and data manipulation (Islam et al., 2023; Shao et al., 2024).

#### 3.3.5.1 Data

We find that training on synthetic data outperforms human-written data, so our approach is heavily weighted towards the use of synthetic examples. We use carefully-curated seed prompts for few-shot generation of novel mathematical problems, and LLM-as-judge techniques to determine the correctness of novel problem-solution pairs. We find that the choices of prompt seeds, correctness validation, and final dataset filtering have a substantial impact on the quality of our reasoning-expert models.

#### 3.3.5.2 Training

**Supervised Fine-Tuning.** For SFT, we leverage synthetic mathematical reasoning datasets that have undergone extensive LLM-driven filtering for solution correctness. We find, similar to Toshniwal et al. (2024), that strict correctness cut-offs are not needed for optimal SFT performance.

**Preference Tuning.** We employ preference tuning following SFT across one of two datasets, dependent on the downstream model training stages: The first dataset is comprised of human-rated preferences on paired completions to reasoning prompts. The second, fully-synthetic dataset comprises correct and incorrect paired solutions to reasoning prompts. We find that, unlike in SFT, solution correctness is of critical importance in preference training (e.g., so that preferences are not accidentally inverted), and in the absence of human-written ratings, we use a mixture of programmatic and model-driven verifiers to evaluate solution correctness.

**Merging.** We find that using candidate models exhibiting maximal reasoning performance is sometimes detrimental to the cross-capability merge under particular merging strategies. We observe that optimal merged performance is achieved when first merging various reasoning-tuned and instruction-tuned expert models, and this yields a sufficiently high-signal proxy for selecting Pareto-optimal candidates to merge with a broader mix of models downstream. We employ this selection for our final set of candidate models, with the exact selection criteria along the Pareto frontier being dependent on downstream merging strategies.

Training hyperparameters for SFT and preference tuning are in Appendix B.4.

### 3.3.6 Long Context

**Data.** Given the complexity of human annotation for long-context tasks, we synthetically generate long-context examples. We sample from our long-context pretraining dataset and prompt Command R+ Refresh to generate question-answer pairs based on randomly selected fragments within 8,192 tokens (Xiong et al., 2024). To ensure high-quality, we use our reward model to select the best generation from a pool of candidates. The selected question-answer pairs are then concatenated to the original samples to construct our synthetic data.

**Training.** We perform one stage of SFT on top of the pretrained model, following a similar approach to our cooldown phase. We use an interleaved training regime with datasets of 16k and 256k sequence lengths at a 3:1 interleaving ratio. Hyperparameters are in Appendix B.5.

### 3.3.7 Safety

AI Safety focuses on quantifying and mitigating harms that can occur from using a given AI model, either to the end user, to the company deploying it, or to society at large. Harms can arise from a single piece of generated content (e.g. hate speech). They can also be distribution-based, which is the case when the model is biased towards certain groups. This section focuses on model safety at the instance level, that is, how we decrease the risks stemming from single generative instances of a given model. We include a distribution-based evaluation (§4.6) and consider it to be a form of robustness (Seshadri & Goldfarb-Tarrant, 2025).

**Cohere's core Safety behaviour.** We focus on practical safety considerations, driven both by model capabilities and deployment use cases. We consider two main settings in which our models can be deployed:

- **The Default setting**, in which the model is used entirely outside of Cohere (e.g. open weights release). In this scenario, we lack control of the preamble structure and deployment context. We ensure that the model behaves according to Cohere's Core Safety behaviour in this general setting.
- **The Enterprise setting**, in which the model is deployed by Cohere to one of our enterprise partners.

Here the safety behaviour of the model is controllable by the preamble, to meet different enterprise needs exceeding Core Safety behaviour. The controllable behaviours are called "Safety modes". There are currently two safety modes; **contextual**, and **strict**.

Our Core Safety behaviour focuses of five key areas where we want to prevent the purposeful propagation of harmful content online: Violence and hate, Misinformation, Self-harm, Child Sexual Exploitation and Abuse (CSEA) and Sexual content. In the default setting, we expect the model to be able to answer requests for information on those topics (covering factual elements such as statistics, educational content); however it should not generate any unsafe content, that is, supporting, encouraging or otherwise enabling harm. In the enterprise setting, the contextual mode is similar, but allows sexual content. The model behaviour can be made stricter by using the strict mode, which prevents the model from covering any topic related to our key focus areas, as well as from generating profanity.

#### *3.3.7.1 Data*

**Pretraining.** We perform two stages of safety-related pretraining filtering: first, we remove known domains for CSEA and sexual content, and second, we use a classifier to remove generally low quality content, including sexual content.

**Post-training.** In post-training, we use both SFT and preference datasets, with a combination of manual and automated labels. Safety annotation is performed by internal annotators and specialist external vendors, who are specifically trained for our Safety concepts and tasks. Our close interaction with internal Safety annotators provides additional benefits due to the potentially distressing nature of the data. We increase the diversity of our post-training data via both LLM 'personas' and LLM-based reformulations. We generate completions corresponding to different styles, identities and belief systems via diverse LLM personas. Additionally, we use our LLM to reformulate content (preserving overall semantics but changing form), thus increasing data diversity and making sure that the preferred completions are consistent with our refusal policy (in particular, the model should not apologise for refusing to generate unsafe content, which creates a common dataset artifact (Chen & Goldfarb-Tarrant, 2025)).

**Balancing safety and refusal behaviour.** Ensuring that the model cannot produce harmful content means that a lot of training data shows refusal as the preferred behaviour. It is crucial to balance such data points with authorised user requests and compliant completions to prevent the model from over-relying on refusal – as previously referred to in the literature as the balance between harmlessness and helpfulness (Bai et al., 2022). The balancing prompts can be split into two sets: user requests which are information requests on safety topics, and benign user requests with similar vocabulary and structure as unsafe prompts.

#### *3.3.7.2 Training*

Improving overall model safety means finding a fine balance between over- and under-refusal. We find it crucial to split datasets in two: namely in their safety-increasing (where the model should refuse) and helpfulness-inducing (where the model should answer) components. This allows us to balance these aspects differently during training. We use both SFT and offline preference tuning. We find offline preference tuning crucial in limiting over-refusal, however, it is less efficient than SFT at making the model safer. We observe this behaviour both on 8B models and 111B models, with the main difference between the two regimes being the effect of regularisation, with larger models more prone to overfitting. Overall, the biggest impact on our model's ability to respond safely and helpfully is achieved in the polishing process described in Section 3.5.

The Safety expert differs from other experts in that during the preference tuning stage we combine an offline preference loss with an equally weighted SFT loss. Preference tuning focuses on reinforcing helpfulness via helpfulness preference pairs, while SFT focuses on reinforcing safety via safety-inducing data. We find that IPO and DPO perform similarly, with SLiC showing a worse trade-off between over- and under-refusal, so we use IPO. Full details on SFT and preference hyperparameters are in Appendix B.6.

## 3.4 Merging

### 3.4.1 Definition

Model merging refers to the process of combining a set of model parameters $\theta_i$ for $i \in [1, K]$, into a single combined model $\theta_{merged} = f(\theta_1, ..., \theta_K)$, where $f(\cdot)$ is some merging function. The merging function can range in complexity from simple averaging (Izmailov et al., 2018; Wortsman et al., 2022; Li et al., 2022) to methods based on Fisher information (Matena & Raffel, 2022) and sign agreement between models (Yadav et al., 2023; Yu et al., 2024). Model merging produces a single set of model parameters, resulting in faster inference than ensembling and lower memory requirements than runtime query routing.

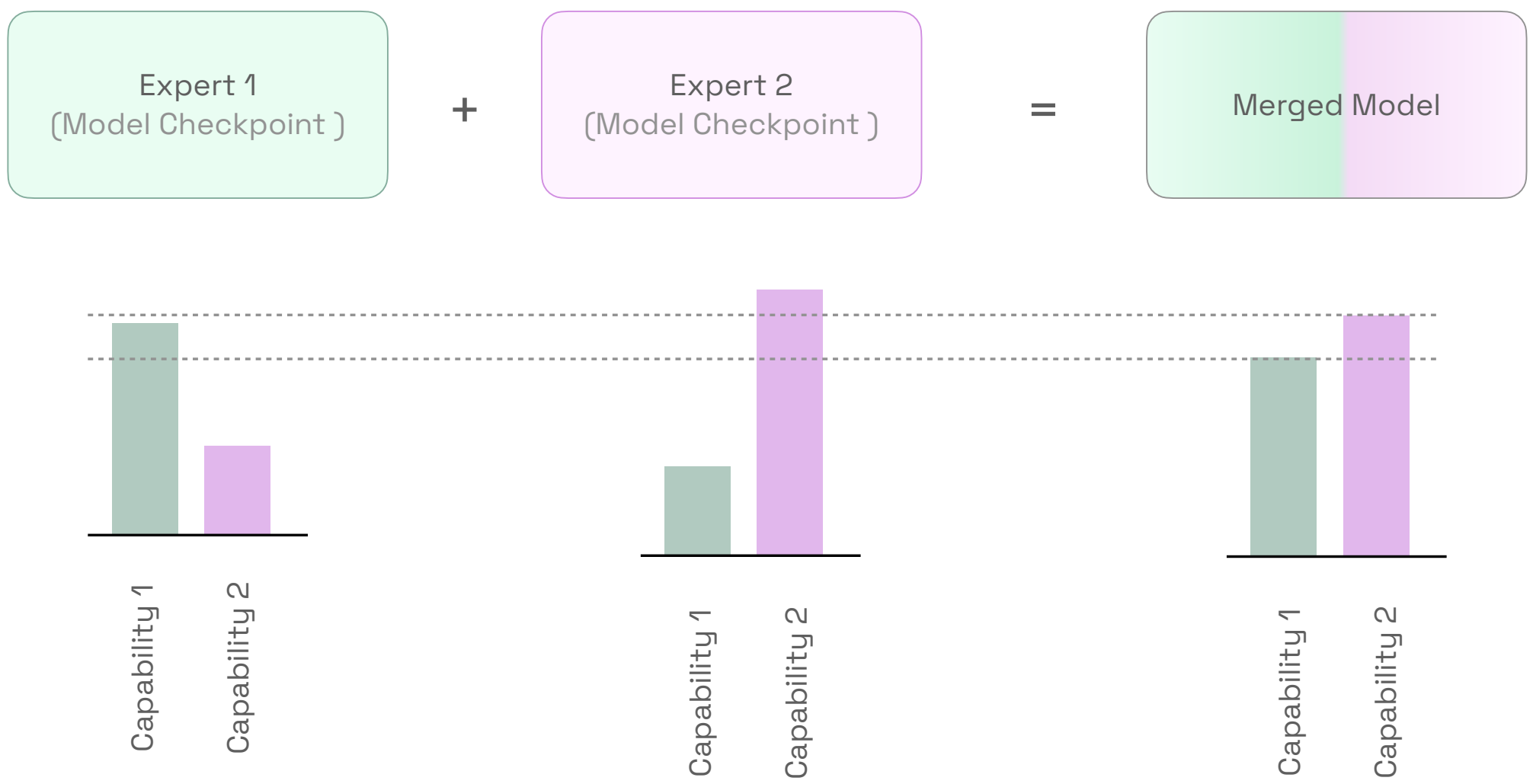

Figure 4: Model merging allows teams to build domain expert models that excel at different capabilities independently. These experts are merged into a single model that retains close-to-expert capability levels across multiple domains or capabilities.

### 3.4.2 Merging Taxonomy

We here list the different merging techniques, each using different models and having different goals.

**Expert merging.** Expert merging refers to the process of combining a set of models with different capabilities to produce a single monolithic model with those capabilities. In this setting, the input models will likely be trained on various datasets and exhibit performance along a single domain only. The aim is to produce a single set of parameters that preserves as much of the individual 'expert' performance as possible. Expert merging is a core feature of the Command A training pipeline, and we describe it in more detail in §3.4.3.

**Merging as Polyak averaging.** Merging may be used to achieve a form of Polyak averaging (Ruppert, 1988; Polyak & Juditsky, 1992). Here, the input models are checkpoints from different points along a single training run, and merging acts as a smoothing operation that reduces the effects of noise inherent in stochastic gradient-based optimisation.

**Seed merging.** Merging may also reduce the effects of random seeds (e.g., for initialization or data loader ordering). Merging the final checkpoints from multiple equivalent training runs with different random seeds can reduce the risk of overfitting and lead to a more robust model.

**Interpolation for capability recovery.** We observe multiple instances of capability forgetting Kirkpatrick et al. (2017), whereby training an expert on one capability degrades performance on other capabilities. This is a particular issue for long-context abilities since experts are generally trained on top of a long-context capable model but with training schemes that use short context lengths. In this situation, merging an expert with the original base model can recover a significant proportion of the original capability while retaining the new expert capability. This setting is closely related to the WARP approach (Ramé et al., 2024).

### 3.4.3 Expert Merging

The overall goal for an enterprise-ready LLM is a single monolithic model, with multiple capabilities. These capabilities can sometimes be orthogonal (e.g., code and safety competencies have very different data distributions) and may involve different scales of training data. For example, it is more straightforward to generate high volumes of synthetic data in more easily verifiable domains, such as code and reasoning, compared to domains like safety or RAG, where human-annotated data is more prevalent. These differences introduce technical and operational challenges: *how can we enable asynchronous development of model capabilities, and jointly optimise for a range of capabilities with highly varied training dynamics?*

Model merging enables multiple teams to work asynchronously on improving different capabilities, with their contributions merged together in parameter space. The capabilities exhibited by Command A cover a wide range of data scales, that would be non-trivial to combine into a single dataset and optimisation strategy. Merging allows each team to separately optimise hyperparameters and data scale for peak performance on their capability of interest. Our final model was informed by 500 separate evaluation metrics, which would have been significantly less practical in a more centralised organisational structure. Merging is computationally cheap, allowing us to quickly and easily rebalance the capabilities of the final model.

We apply merging at two points in the overall training pipeline: firstly, to combine a set of expert models trained using SFT into an initial 'SFT model soup'; secondly, to combine a set of experts that were trained using offline preference optimisation techniques on top of the SFT soup, giving an 'off-pref soup'. At both stages, our aim is to jointly maintain as high a proportion of the expert capability as possible while also allowing for rebalancing of the overall capabilities of the final model.

#### *3.4.3.1 Linear merging is simple but effective*

We employ linear merging (also known as weight averaging), with weights chosen by manual search. We find that, broadly speaking, the interaction between expert weights and resulting model performance is fairly intuitive; increasing the weight of a domain expert is likely to increase the performance in that domain. However, this relationship is not perfect, and the corresponding degradation in performance of other (implicitly downweighted[2]) domains is much less predictable. We therefore search across merging weights using a combination of heuristics (i.e., upweight experts for domains in which a merge candidate is underperforming) and brute force search (i.e., perturb the weights for each expert, centred around the current candidate). We experimented with more complex merging methods (e.g., SLERP (Ramé et al., 2024) and task vectors (Ilharco et al., 2023)) but found no significant performance improvements, at the cost of increased complexity. In addition, linear merging is associative, meaning that a linear merge of linear merges can be expressed as a single merge operation, improving the interpretability of a complex training pipeline.

#### *3.4.3.2 Consistency is more important than optimality*

All expert models are initialised from a common 'general instruction following' model, for two reasons. Firstly, some domain experts make use of special tokens (e.g., tool calls) whose embeddings otherwise remain untrained. We find that selectively merging these embeddings only from checkpoints where they are trained is beneficial, but suboptimal. Using a shared generalised instruction-following model as initialisation for each expert and merging the special token embeddings as normal performs much better, even though these embeddings are likely to be lower quality. Secondly, we find that the post-training process generally degrades long-context performance, and that this is challenging to recover. Starting from a generalised model that is 'long-context capable' preserves long-context performance more easily throughout the training pipeline.

We find it valuable to include 'leave-one-out' merges as part of the search process, to reveal instances where one expert model causes performance degradation of others, or 'collisions'. To address this, we include a small amount of cross-domain data in each expert's training, to act as a regulariser and ensure that each expert remains 'compatible' with the other experts. We also observe that collisions can be caused by small inconsistencies in the style or formatting of the common data between experts. In combination this implies that maintaining some consistency between expert models is more important than absolute expert performance.

[2]For linear merging, the weights must sum to 1. Increasing the weight of one expert therefore requires reducing the weight of one or more of the other experts.

#### 3.4.3.3 *Merging is cheap, evaluation is expensive*

We note that most prior work on model merging assumes that the set of input experts is fixed, and seeks to find a single merging method that optimises some value function, generally a single metric or small number of metrics (e.g., Wang et al., 2024a). These methods often involve a large number of hyperparameters (Ilharco et al., 2023) or extensive search over merge weights (Khalifa et al., 2025). By contrast, our goal is to optimise for a wide range of capabilities and many metrics. This introduces a further challenge generally not acknowledged by the literature; while model merging is cheap and fast, evaluating each merge requires significant inference time and compute. Evaluation is therefore a significant bottleneck when applying model merging in a production context. The set of input models is also not fixed, and a significant portion of the effort towards successful merging involves making changes to the training scheme used by the experts.

## 3.5 Polishing

Model merging provides a powerful mechanism for combining a diverse set of experts into a single model. However, combining experts trained to target specific capabilities does not guarantee the final model's alignment with human preferences. To address this, we introduce a polishing phase as the final post-training step. This phase serves two critical purposes: fixing any artifacts introduced during model merging and aligning the final model with human preferences.

Unlike other specific capabilities such as coding or instruction-following, human alignment has a cross-domain effect and influences every aspect of the model's behaviour. The polishing phase ensures that the model adheres to human expectations, including tone and style, without sacrificing technical competence.

Polishing is divided into three steps. First, we apply SFT on a subset of our highest quality datasets. Second, we apply offline preference tuning, and finally, we apply online Reinforcement Learning from Human Feedback (RLHF). We find that ping-pong-style interleaving of offline and online preference learning helps improve alignment with human preferences while avoiding regressions and mitigating reward model hacking.

**Supervised Fine-Tuning (SFT).** We employ a best-of-N SFT approach (Stiennon et al., 2020) where we synthetically generate four candidate completions for each prompt. We leverage our reward model (§3.2.3) trained on human preference data to rank these completions. We then apply SFT using the highest-ranked completions, ensuring that the model learns from the most highly rewarded responses.

**Preference Tuning.** We use offline preference training to align our model with human preferences. We select completions with the highest reward scores as preferred completions, and use the completions with the lowest reward scores as dis-preferred. Additionally, we refine the dataset by filtering out prompts exhibiting a low average reward. To further improve the model's proficiency in instruction-following, mathematical reasoning, and coding, we incorporate domain-specific preference data into our training mixture. For instruction-following, completions that correctly adhere to all instructions are considered preferred, while completions failing to meet all instruction criterion are labelled as dis-preferred. To construct preference data for mathematical reasoning, we categorise completions that yield correct answers as preferred and those failing to produce accurate solutions as dis-preferred. Similarly, for code generation tasks, code snippets passing all unit tests serve as preferred completions, while those failing the tests are used as dis-preferred completions. We also filter these preference datasets by removing samples for which the preferred completions are assigned a lower score than the dis-preferred completions by our reward model. We rely again on the SRPO loss due to its robustness and its self-refinement abilities (§3.2.2.1). In our implementation of SRPO, following Grinsztajn et al. (2024), we average the log-likelihoods of preferred and dispreferred completions to control for variations in the completion length.

**Reinforcement Learning from Human Feedback (RLHF).** To enhance the alignment of our model with human preferences, we further employ Reinforcement Learning from Human Feedback (RLHF). We use online CoPG (§3.2.2.2) with two generations per prompt. The prompts used for RLHF training are derived from a subset of those previously used during SFT, including reasoning, multilingual, coding, and preference-based tasks prompts. We regularize training using an auxiliary $L_2$ loss with the reference policy, and an SFT loss using a high-quality subset of post-training data.

| Area | Benchmarks |
|---|---|
| Academic, General Knowledge and Instruction Following (§4.1) | MMLU; MMLU-Pro; GPQA; IFEval; InFoBench |
| Agents and Tool-Use (§4.2) | TauBench; BFCL. |
| Multilingual (§4.3) | MMMLU; NTREX; FLoReS; MGSM; mArenaHard (LLM-as-a-Judge); Language Confusion Benchmark; Al-Qasida; INCLUDE 44; mTauBench. |
| Code (§4.4) | LBPP; HumanEvalPack; MBPP+; Spider; Bird SQL; RepoQA; LiveCodeBench; BigCodeBench; SWE-Bench Diff Generation; Aider Polyglot; internal datasets. |
| Math and Reasoning (§4.5) | MATH; AIME; LiveBenchMath; Waterloo; OpenQuant; FinanceBench; OmniMath. |
| Safety (§4.6) | XSTest; internal datasets. |
| Long-Context (§4.8) | Needle-in-a-Haystack; RULER; RulerQA. |

Table 2: Benchmark datasets used to evaluate Command A models, grouped by area.

# 4 Results

We report results from a diverse and extensive set of evaluations benchmarking the performance of Command A and Command R7B. We evaluate a broad range of capabilities using public academic datasets and internal evaluations. Table 2 gives an overview of the capability areas we focus on and the corresponding benchmarks. We present a snapshot of results on a representative subset of these evaluations in Table 1 opening this report. Full details for each dataset are available in the corresponding sections.

We compare our models against open and closed models in similar parameter count ranges. Wherever possible, we show externally reported results with comparable evaluation settings. Where these are not available, we attempt to internally reproduce these results as faithfully as possible given the information provided publicly.

## 4.1 Standard Benchmarks

While our primary aim is to build a highly performant model for enterprise use cases (§4.7), we also measure performance on standard academic datasets to evaluate baseline model knowledge and capabilities. Where applicable (MMLU, MMLU-Pro, GPQA), we follow the `simple-evals` implementation, including data, task settings, prompting, and answer parsing. More details can be found in Appendix B.7.

| Model | MMLU | MMLU-Pro | GPQA | IFEval | InFoBench |
|---|---|---|---|---|---|
| Command A | 85.5 | 69.6 | 50.8 | 90.9 | 94.9 |
| GPT-4o | 89.2 | 77.9 | 53.6 | 83.8 | 94.0 |
| DeepSeek V3 | 88.5 | 75.9 | 59.1 | 86.1* | 94.3 |
| Llama 3.3 70B Instruct | 86.0 | 66.0 | 50.5 | 92.1 | 92.8 |
| Llama 3.1 405B Instruct | 88.6 | 73.0 | 49.0 | 88.6 | 93.9 |
| Mistral Large 2 | 85.2 | 67.9 | 48.6 | 83.8 | 93.3 |
| Claude 3.5 Sonnet | 89.5 | 78.0 | 65.0 | 90.2 | 93.9 |
| Gemini 2.0 Pro | 89.3 | 79.1 | 64.7 | 87.3 | 92.2 |
| Command R7B | 65.2 | 42.4 | 26.3 | 77.9 | 85.6 |
| Llama 3.1 8B Instruct | 71.1 | 46.5 | 23.4 | 78.6 | 90.1 |
| Ministral 8B | 71.1 | 43.0 | 23.4 | 59.0 | 88.3 |
| Gemma 2 9B Instruct | 73.5 | 50.6 | 31.3 | 74.4 | 87.2 |
| Gemini 1.5 Flash-8B | 74.8 | 48.4 | 31.6 | 88.0 | 88.3 |

Table 3: Results for Command A and Command R7B on standard academic benchmarks. *Note that for IFEval, Liu et al. (2024a) report only the prompt-level strict accuracy. We report the average of the prompt- and instruction-level strict accuracies for all other models (see Appendix B.7).

We note that academic benchmarks have various limitations such as saturation, bias and alignment to real-world performance (Kiela et al., 2021). Human assessment of model capabilities can be undesirably influenced by confounders (Hosking et al., 2024), be subject to idiosyncratic, conversational and demographic variance (Kirk et al., 2024), and demonstrates imperfect correlation to academic benchmarks (Schaeffer et al., 2025). Enterprise-relevant capabilities are often not well-represented in these benchmarks, so we augment our evaluations with enterprise-oriented signal (e.g. §4.2, §4.7), and human annotation based evaluation (§4.11).

Table 3 shows results on these selected benchmarks. Command A is competitive across all benchmarks, generally outperforming similarly-sized models while remaining competitive with considerably larger and less-efficient models. On the instruction-following benchmarks, we observe that Command A performs competitively across both IFEval and InFoBench. Specifically, it outperforms all similarly sized models on InFoBench and is outperformed only by Llama 3.3 70B Instruct on IFEval. We also note that Command A represents a substantial improvement over our previous Command R+ Refresh model.

## 4.2 Agentic Tool Use

| Model | ChatRAGBench | StrategyQA | Bamboogle | DROP | HotPotQA |
|---|---|---|---|---|---|
| Command A | 72.9 | 76.7 | 76.0 | 91.1 | 92.1 |
| GPT-4o | 66.6 | 81.2 | 76.0 | 89.5 | 92.1 |
| DeepSeek V3 | 40.3 | 73.8 | 70.4 | 85.7 | 90.1 |

Table 4: **Standard RAG evaluations**. Correctness is determined following the procedure in Verga et al. (2024) where a panel of LLMs judges the model's generation against a reference answer.

| Model | BFCL Overall | Live AST | Multi-turn |
|---|---|---|---|
| Command A | 63.8 | 80.5 | 25.5 |
| Llama 3.3 70B Instruct | 51.4 | 62.8 | 6.9 |
| Mistral Large 2 | 58.5 | 69.9 | 23.8 |
| Qwen 2.5 72B Instruct | 63.5 | 79.0 | 24.6 |
| Claude 3.5 Sonnet | 56.5 | 78.9 | 41.0 |
| Claude 3.7 Sonnet | 58.3 | 78.4 | 48.4 |
| DeepSeek V3 | 58.6 | 68.4 | 18.6 |
| GPT-4o | 72.1 | 79.8 | 47.6 |
| Command R7B | 52.2 | 69.2 | 5.0 |
| Llama 3.1 8B Instruct | 50.9 | 61.1 | 9.6 |
| Gemma 2 9B Instruct | 51.6 | 68.0 | 1.6 |
| Ministral 8B | 51.8 | 64.9 | 11.4 |
| Qwen 2.5 7B Instruct | 53.7 | 67.4 | 7.6 |

Table 5: **BFCL Results**. All numbers taken from official leaderboard. Where leaderboard entries exist for both function calling and prompted, we take the larger of the two reported values.

| Model | Taubench Retail | | | | Taubench Airline | | | |
|---|---|---|---|---|---|---|---|---|
| | P@1 | P@2 | P@3 | P@4 | P@1 | P@2 | P@3 | P@4 |
| Command A | 60.0 | 49.8 | 44.1 | 40.4 | 45.3 | 36.9 | 32.2 | 29.0 |
| Llama 3.3 70B Instruct | 6.2 | 5.7 | 5.49 | 5.3 | 35.3 | 33.6 | 32.4 | 31.5 |
| Mistral Large 2 | 53.3 | 37.8 | 29.0 | 23.1 | 27.2 | 14.2 | 9.4 | 7.1 |
| Llama 3.1 405B Instruct | 29.1 | 17.5 | 12.8 | 10.4 | 26.0 | 17.3 | 13.5 | 12.0 |
| DeepSeek V3 | 54.8 | 41.2 | 34.1 | 30.4 | 25.5 | 14.0 | 12.0 | 12.0 |
| GPT-4o | 60.6 | 49.0 | 42.4 | 37.7 | 43.0 | 31.8 | 26.3 | 22.3 |
| Claude 3.5 Sonnet | 69.2 | 57.6 | 50.9 | 46.2 | 46.0 | 32.6 | 26.3 | 22.5 |

Table 6: **Taubench Results.** We follow the original experimental setup from Yao et al. (2024). Pass@k (P@k) evaluates a model's consistency; for example, Pass@4 is the probability that a model answers the same question correctly 4 times. Scores are aggregated over 10 runs.

.

**Standard RAG Benchmarks.** We evaluate on several RAG benchmarks that test the model's ability to answer questions conditioned on source documents. In DROP (Dua et al., 2019) and HotPotQA-distractor (Yang et al., 2018), the model is given a question and set of pre-retrieved relevant documents. Bamboogle (Press et al., 2022) and StrategyQA (Geva et al., 2021) are multi-hop question answering datasets where models must submit one or more sequential or parallel queries to a search engine to gather documents and arrive at the answer. Finally, we show results averaged over the ten datasets in ChatRAGBench (Liu et al., 2024d) that cover a variety of domains situated in a multi-turn conversation. Results are shown in Table 4.

**Berkeley Function-Calling Leaderboard (BFCL).** BFCL is one of the most widely used evaluations of LLM tool use / function calling capabilities and maintains an independently run leaderboard (Yan et al., 2024). Evaluations include simple single step tool calls, measures of tool irrelevance, and a multi-turn subset which simulates much harder scenarios over long action trajectories. Results are shown in Table 5.

**Taubench.** Taubench is a complex agentic tool-use benchmark that simulates a customer support agent in two settings: airline and retail (Yao et al., 2024). The agent model has access to a set of tools for reading and writing to a provided database and must help a simulated user in accomplishing a given task such as changing flight or returning a product order. Results are shown in Table 6.

## 4.3 Multilingual

Command A supports 23 key languages of global business: English, French, Spanish, Italian, German, Portuguese, Japanese, Korean, Arabic, Chinese, Russian, Polish, Turkish, Vietnamese, Dutch, Czech, Indonesian, Ukrainian, Romanian, Greek, Hindi, Hebrew, and Persian. We evaluate performance on many of these languages (and beyond) on both academic and internal enterprise focused benchmarks, as well as public benchmarks important for business use such as language consistency and steerability, and dialect awareness.

We assess the general multilingual capability of Command A through machine translation via NTREX-128 (Federmann et al., 2022), which contains human translated news domain documents, FLORES-200 (Team et al., 2022; Goyal et al., 2022); and multilingual mathematical reasoning (MGSM; Shi et al. (2022)). We further evaluate Command A's understanding of regional contexts through INCLUDE (Romanou et al., 2025), a large-scale region-specific evaluation suite in 44 languages.

Results for machine translation on NTREX are shown in Table 7. We use the COMET-20 metric (Rei et al., 2020), one of the top performing MT metrics (Freitag et al., 2023). Rather than mark single winning models, we mark winning clusters of models by taking into account the effect size of the metric. A model is in a winning cluster if its score difference to the best model is smaller than 1.67 points. This threshold equates to 75% agreement with humans (Kocmi et al., 2024): humans will agree with automatic metric on 3 out of 4 pairwise system comparisons that have a difference of 1.67 COMET-20. Further academic results (MGSM and INCLUDE-44) are in Appendix B.2.

To evaluate more general and diverse capabilities, we ran an LLM-as-judge arena-like evaluation of responses to mArenaHard, a dataset of 500 challenging queries from Chatbot Arena, originally in English, translated into 23 languages (Dang et al., 2024). As shown in Table 8, Command A is preferred across all 23 languages versus Llama 3.3 70B Instruct, Llama 3.1 405B Instruct, and DeepSeek V3.

We also conduct a human-annotated arena-like evaluation. Figure 5 shows the results of an internal evaluation set consisting of 100 translated prompts [3] from English that focus on instruction-following ability. Command A performs favourably in multilingual head-to-head human evaluations against comparable models across 9 priority languages. Command A outperforms the Llama 3.3 70B Instruct and Llama 3.1 405B Instruct across all evaluated languages. Versus DeepSeek V3 and Mistral Large 2, Command A is favoured across 8 of 9 languages. Notably, Command A is favoured in Chinese compared to DeepSeek V3 and is favoured in French compared to Mistral Large 2. It is also favoured against GPT-4o on Arabic and Korean, and is competitive on Spanish, German, Italian, and Chinese.

[3]Models commonly had issues generating completions for one of the prompts across different languages, the win/tie/loss rates for these are based on 99 prompt-completion pairs

|  | ar | cs | de | el | es | fa | fr | he | hi | id | it | ja | ko | nl | pl | pt | ro | ru | tr | uk | vi | zh | Avg. | FLORES |
|---|---|---|---|---|---|---|---|---|---|---|---|---|---|---|---|---|---|---|---|---|---|---|---|---|
| Command A | **60.0** | 84.3 | **60.2** | 79.5 | **74.2** | **56.3** | **67.2** | 66.2 | 67.8 | 78.7 | 74.2 | 60.6 | 66.8 | **63.2** | 73.3 | 74.7 | 74.1 | 64.2 | 80.1 | 68.3 | 63.7 | 56.3 | 68.8 | 81.2 |
| GPT-4o | **60.8** | **85.9** | **61.3** | **83.0** | **74.7** | **57.9** | **68.6** | **70.8** | **70.7** | **82.2** | **76.1** | **64.1** | **68.9** | **64.5** | **75.4** | **76.8** | **76.8** | **65.0** | **83.9** | **70.7** | **66.1** | **56.9** | **71.0** | **83.0** |
| Gemini 2.0 Flash | 58.9 | **85.1** | **61.7** | **82.2** | **75.0** | **57.6** | **67.9** | 67.3 | **71.5** | **81.6** | **75.4** | **62.9** | 66.9 | **64.6** | **75.2** | **75.8** | **76.8** | **65.9** | **84.8** | **71.5** | **65.9** | 56.4 | **70.5** | **82.8** |
| Gemini 1.5 Pro | 58.4 | **85.3** | **61.3** | **83.1** | **74.1** | **57.2** | **68.0** | 68.8 | **71.4** | **81.2** | **75.9** | 61.6 | 65.2 | **63.8** | **75.7** | **76.4** | **77.6** | **65.7** | **84.2** | **72.2** | **66.0** | 56.5 | **70.4** | **82.8** |
| Claude 3.7 Sonnet | **59.5** | **86.1** | **61.5** | 80.7 | 73.0 | 55.9 | **67.3** | **70.2** | 69.6 | **81.2** | **75.7** | **63.6** | **69.0** | **64.0** | **75.3** | **75.3** | 75.4 | **65.5** | 82.6 | **71.6** | **65.5** | 56.8 | **70.2** | **82.7** |
| DeepSeek V3 | **59.8** | **85.0** | **61.0** | 76.7 | 73.3 | 55.3 | **67.6** | 68.3 | **70.9** | **81.6** | **74.8** | **64.4** | **68.2** | **63.8** | **74.3** | **76.0** | 74.6 | **64.9** | 83.0 | 69.0 | **65.8** | **58.5** | **69.8** | 76.3 |
| Llama 3.1 405B Instruct | 52.0 | 81.3 | 59.0 | 71.5 | 71.4 | 49.1 | 64.3 | 63.9 | 64.8 | 78.5 | 73.3 | 59.8 | 63.3 | 62.9 | 70.3 | 72.3 | 73.2 | 60.9 | 77.7 | 62.7 | 60.2 | 54.3 | 65.8 | 79.1 |
| Mistral Large 2 | 52.1 | 77.7 | 59.4 | 69.7 | 71.4 | 45.5 | 65.7 | 63.2 | 61.3 | 73.3 | 73.8 | 58.9 | 63.2 | 59.2 | 67.9 | 74.1 | 69.7 | 60.7 | 68.5 | 63.8 | 58.3 | 53.2 | 64.1 | 77.1 |
| Llama 3.3 70B Instruct | 45.2 | 76.1 | 57.6 | 64.7 | 69.1 | 43.6 | 62.7 | 59.9 | 63.4 | 75.6 | 70.7 | 57.0 | 57.3 | 61.4 | 67.7 | 70.9 | 70.1 | 58.1 | 70.5 | 61.2 | 58.3 | 53.5 | 62.5 | 75.8 |
| Qwen 2.5 72B Instruct Turbo | 48.3 | 70.8 | 56.0 | 41.8 | 70.2 | 28.8 | 63.4 | 19.0 | 49.6 | 74.3 | 69.5 | 57.7 | 51.8 | 57.4 | 60.2 | 73.6 | 56.4 | 58.3 | 65.0 | 49.6 | 55.9 | 54.5 | 56.0 | 69.8 |
| Command R7B | **50.6** | 61.8 | 54.7 | 30.2 | 69.2 | 32.8 | 61.4 | -15.1 | 35.9 | 58.5 | 68.4 | 50.7 | 52.2 | 48.4 | 46.6 | 69.1 | 58.4 | 39.8 | 50.1 | 40.7 | 46.0 | 47.6 | 48.1 | 58.6 |
| Gemini 1.5 Flash-8B | 47.5 | 76.0 | **57.1** | **73.0** | **71.3** | **47.7** | **64.3** | **51.6** | **64.8** | **78.0** | **71.2** | **56.1** | 60.2 | **59.5** | **67.6** | **72.3** | **70.2** | **59.8** | **75.1** | **63.6** | **61.5** | **50.9** | **63.6** | **77.0** |
| Claude 3 Haiku | 46.4 | **79.2** | **56.8** | 62.9 | 68.4 | 44.2 | 62.1 | **50.1** | 59.0 | **76.9** | **69.7** | 53.9 | **63.7** | **59.2** | **66.5** | **70.7** | 67.9 | **59.5** | **73.9** | **63.7** | 58.6 | **51.6** | **62.0** | **76.3** |
| Gemma 2 9B Instruct | 42.2 | 73.0 | **56.3** | 61.4 | **70.2** | 40.3 | 62.4 | 36.0 | 60.0 | 76.3 | **69.9** | **54.5** | 53.8 | 57.2 | **66.3** | **72.0** | 66.4 | 56.9 | 70.4 | 60.2 | 57.5 | **51.0** | 59.7 | 72.6 |
| Llama 3.1 8B Instruct | 26.7 | 61.2 | 50.1 | 40.8 | 63.8 | 27.1 | 54.7 | 30.2 | 46.9 | 67.7 | 63.9 | 42.5 | 41.0 | 52.6 | 53.6 | 64.3 | 57.5 | 47.0 | 48.8 | 47.0 | 51.1 | 44.6 | 49.2 | 63.8 |
| Ministral 8B | -12.2 | 42.5 | 52.0 | 34.9 | 65.4 | -10.3 | 56.9 | 9.2 | 18.0 | 55.5 | 65.2 | 38.2 | 34.6 | 44.8 | 31.4 | 67.4 | 34.5 | 48.1 | 23.3 | 38.7 | 33.6 | 38.6 | 36.8 | 52.3 |
| Qwen 2.5 7B Instruct Turbo | 2.7 | 29.5 | 39.0 | -37.5 | 59.6 | -34.5 | 49.5 | -68.4 | -12.9 | 56.1 | 54.4 | 32.7 | 11.3 | 38.0 | 20.1 | 61.2 | 15.3 | 26.2 | 17.1 | -5.3 | 39.1 | 47.0 | 20.0 | 33.7 |

Table 7: **Machine translation (COMET-20) scores on NTREX.** Average over FLORES (COMET-20) is also shown. System scores in the winning cluster are bold per language.

|  | Avg. | ar | cs | de | el | en | es | fa | fr | he | hi | id | it | ja | ko | nl | pl | pt | ro | ru | tr | uk | vi | zh |
|---|---|---|---|---|---|---|---|---|---|---|---|---|---|---|---|---|---|---|---|---|---|---|---|---|
| **Command A** vs. Llama 3.3 70B Instruct | **78.9** | 81.2 | 77.3 | 81.4 | 78.7 | 72.8 | 76.4 | 81.6 | 76.1 | 84.0 | 81.9 | 81.6 | 74.9 | 84.9 | 84.6 | 78.8 | 79.9 | 76.7 | 77.0 | 69.7 | 79.0 | 80.0 | 82.6 | 74.4 |
| **Command A** vs. Llama 3.1 405B Instruct | **78.7** | 80.4 | 77.4 | 81.2 | 78.7 | 73.6 | 82.7 | 78.4 | 79.3 | 85.2 | 81.1 | 81.7 | 81.2 | 82.7 | 79.3 | 76.3 | 77.6 | 82.5 | 78.0 | 73.6 | 70.6 | 72.1 | 78.2 | 78.8 |
| **Command A** vs. Mistral Large 2 | **65.9** | 68.4 | 64.8 | 64.4 | 68.5 | 64.1 | 63.8 | 68.6 | 61.0 | 68.0 | 65.6 | 66.8 | 62.9 | 66.6 | 71.2 | 64.8 | 64.9 | 61.7 | 65.5 | 64.2 | 67.6 | 64.5 | 70.2 | 66.9 |
| **Command A** vs. DeepSeek V3 | **53.4** | 54.5 | 54.5 | 55.9 | 54.5 | 52.5 | 52.7 | 52.7 | 53.7 | 54.0 | 51.6 | 53.9 | 51.8 | 56.4 | 53.2 | 52.7 | 50.2 | 54.2 | 51.9 | 52.0 | 50.4 | 55.2 | 55.7 | 54.7 |

Table 8: Command A mArenaHard winrates on 23 languages against open-weights models.

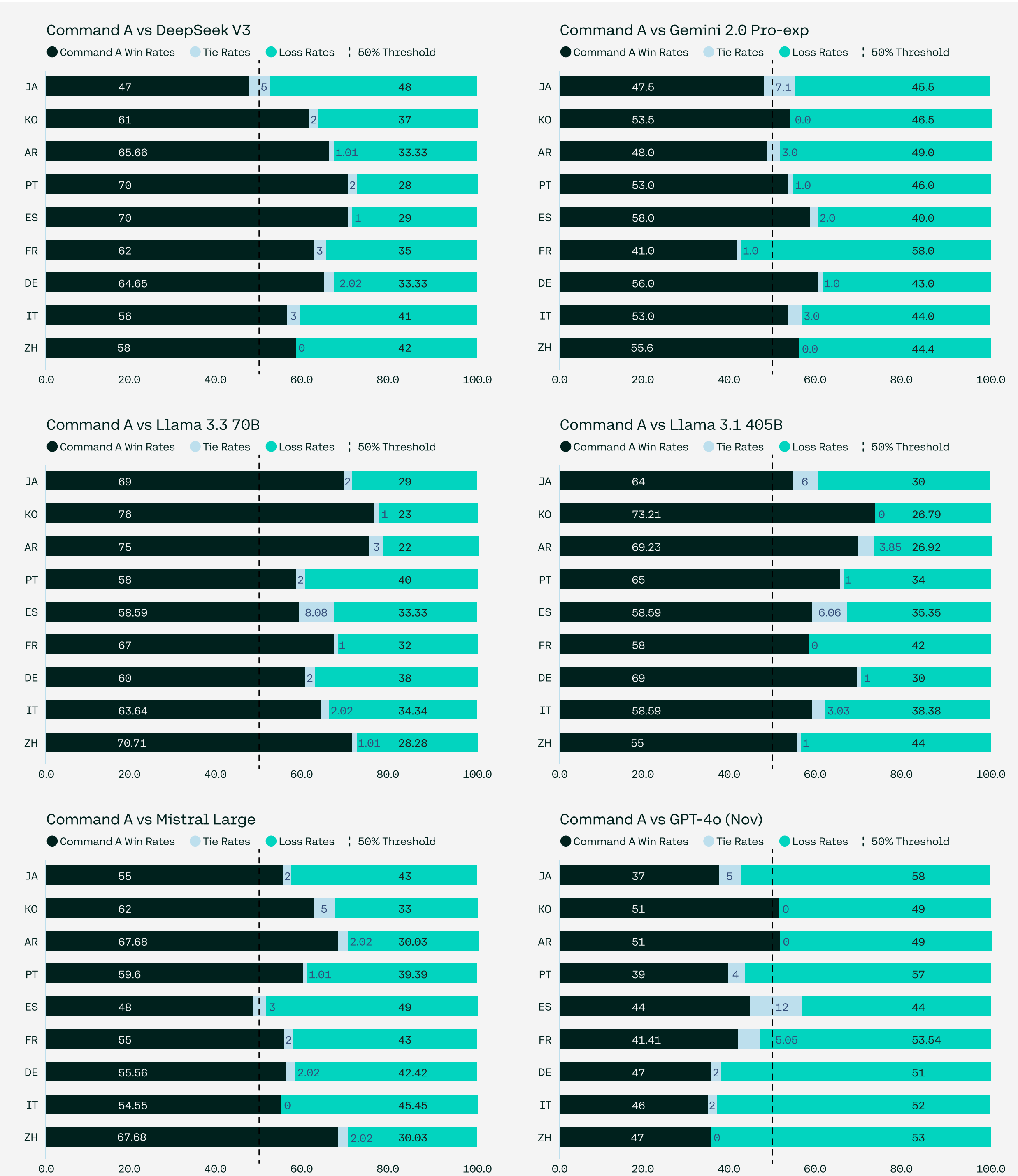

Figure 5: Head-to-head human evaluations against comparable models.

| | MTaubench Retail | | | | | | MTaubench Airline | | | | | |
|---|---|---|---|---|---|---|---|---|---|---|---|---|
| | Avg. | en | fr | ar | ja | ko | Avg. | en | fr | ar | ja | ko |
| Command A | 34.3 | **60.0** | 36.5 | 28.5 | 24.4 | 22.0 | 43.4 | **45.3** | **52.7** | 47.3 | 38.0 | 33.8 |
| GPT-4o | **37.3** | 59.7 | **41.7** | **29.6** | **28.7** | **26.7** | 45.2 | 47.3 | 38.7 | **50.7** | 41.3 | **48.0** |
| Gemini 1.5 Pro | 26.4 | 49.0 | 28.4 | 20.3 | 16.2 | 18.8 | 41.4 | 31.7 | 36.9 | 46.0 | **47.6** | 45.0 |
| Gemini 2.0 Flash | 26.0 | 44.1 | 29.6 | 20.6 | 17.1 | 18.8 | 33.7 | 35.3 | 36.0 | 34.0 | 29.3 | 34.0 |
| Mistral Large 2 | 25.8 | 54.1 | 30.0 | 18.7 | 11.8 | 14.4 | 27.0 | 28.6 | 32.0 | 30.9 | 21.6 | 21.9 |

Table 9: **Multilingual Taubench Results**: We follow the original experimental setup from Yao et al. (2024). We report the per language pass@1 (P@1) score. Scores aggregated over 3 runs.

| | Avg | ar | de | es | fr | hi | id | it | ja | ko | pt | ru | tr | vi | zh |
|---|---|---|---|---|---|---|---|---|---|---|---|---|---|---|---|
| Command A | 93.0 | **98.2** | 95.5 | 94.8 | 93.5 | 94.6 | 84.2 | 94.9 | 93.2 | 93.4 | 89.6 | 93.7 | 93.6 | 92.2 | 90.3 |
| Command R+ Refresh | **95.5** | 94.8 | **98.2** | **97.2** | **97.2** | 96.5 | 89.0 | **97.5** | **96.2** | **94.7** | **91.6** | **96.9** | **97.8** | **98.3** | **90.6** |
| Qwen 2.5 72B Instruct Turbo | 93.0 | 96.4 | 94.0 | 94.3 | 93.7 | 94.1 | 86.6 | 94.0 | 91.3 | 93.0 | 87.9 | 95.8 | 95.4 | 95.7 | 89.2 |
| Claude 3.7 Sonnet | 91.8 | 94.5 | 95.5 | 93.2 | 94.2 | 93.6 | 78.9 | 94.0 | 93.6 | 93.9 | 84.1 | 95.9 | 92.6 | 95.3 | 86.2 |
| Llama 3.3 70B Instruct | 91.3 | 90.5 | 94.7 | 92.9 | 93.0 | **98.2** | **92.1** | 93.3 | 83.3 | 78.9 | 91.2 | 95.0 | 92.9 | 95.9 | 86.5 |
| Gemini 1.5 Pro | 90.6 | 91.7 | 94.4 | 94.5 | 92.0 | 93.5 | 82.9 | 93.3 | 86.1 | 90.6 | 86.7 | 93.4 | 94.8 | 95.6 | 79.1 |
| DeepSeek V3 | 90.6 | 94.9 | 93.8 | 94.2 | 93.5 | 92.2 | 82.6 | 92.8 | 85.5 | 91.5 | 85.3 | 92.9 | 92.3 | 91.8 | 84.3 |
| GPT-4o | 88.9 | 92.2 | 91.0 | 94.9 | 91.7 | 91.9 | 80.9 | 90.4 | 85.3 | 87.4 | 87.0 | 87.9 | 90.7 | 88.0 | 85.5 |
| Mistral Large 2 | 75.9 | 85.3 | 64.7 | 83.4 | 78.3 | 86.9 | 65.0 | 69.6 | 82.6 | 76.1 | 75.2 | 75.6 | 69.3 | 73.1 | 77.2 |

Table 10: Crosslingual line-level pass rate (LPR) from the Language Confusion Benchmark (Marchisio et al., 2024). Models are prompted in English with an instruction to reply in a different language. LPR measures the percentage of answers with all lines in the requested language.

| | Monolingual | Crosslingual |
|---|---|---|
| Command A | **24.2** | **33.5** |
| Gemini 1.5 Pro | 19.3 | 26.4 |
| GPT-4o | 15.8 | 24.7 |
| Claude 3.7 Sonnet | 8.5 | 23.1 |
| DeepSeek V3 | 15.7 | 15.7 |
| Llama 3.3 70B Instruct | 15.2 | 8.3 |
| Qwen 2.5 72B Instruct Turbo | 9.9 | 9.6 |
| Mistral Large 2 | 6.9 | 7.9 |
| Command R+ Refresh | 1.9 | 6.1 |

Table 11: ADI2 score over monolingual and crosslingual prompts in 4 Arabic dialects (Egyptian, Saudi, Syrian, Moroccan) from Robinson et al. (2024). Higher scores indicate greater desired dialect adherence.

Beyond instruction-following, agentic capabilities are important for enterprise use. We evaluate Command A on our own human translated version of $\tau$-bench (Yao et al., 2024).[4] As shown in Table 9, Command A outperforms other widely-adopted LLMs agentic solutions such as Mistral Large 2 and Gemini 1.5 Pro, while being competitive with GPT-4o.

The Language Confusion Benchmark (Marchisio et al., 2024) measures a model's ability to appropriately respond in the desired language of the user. In Table 10, we measure line-level pass-rate (LPR) on crosslingual prompts. Concretely, models are prompted with an English request and an instruction to reply in another language. LPR is the percentage of responses where all lines were in the user's desired language. Command A and its predecessor, Command R+ Refresh, perform very strongly across languages, with the highest and second highest aggregate scores.

We measure Command A's sensitivity to regional dialect in Table 11, which shows ADI2 scores over monolingual and crosslingual prompts in 4 Arabic dialects (Egyptian, Saudi, Syrian, Moroccan) from Robinson et al. (2024). Higher scores indicate more adherence to the desired Arabic dialect. We observe that Command

[4]The number may differ slightly from the official implementation due to extensions for our multilingual evaluation pipeline.

A strongly outperforms comparison models in its ability to adhere to dialect.

## 4.4 Code

We evaluate the code capabilities of Command A across **code understanding**, **code editing**, and **SQL generation** benchmarks.

| | Python | | | Multi-language | | COBOL | | RepoQA |
|---|---|---|---|---|---|---|---|---|
| | MBPP+ | LiveCodeBench | BigCodeBench | LBPP(All) | HE(All) | HE | →Python | |
| Command A | 86.2 | 26.9 | 45.4 | 51.5 | 76.2 | 25.3 | 55.7 | **92.6** |
| Command A Expert | 87.0 | 24.9 | 47.4 | 50.8 | 77.5 | **29.8** | **64.6** | 91.8 |
| Command A Agentic | — | **32.9** | **59.7*** | **65.4** | — | — | — | — |
| Command R7B | 72.0 | 9.0 | 30.9 | 21.9 | 50.7 | 7.0 | 35.4 | 69.6 |
| Command R Refresh | 74.3 | 11.0 | 34.3 | 24.7 | 54.7 | 1.9 | 34.2 | 73.2 |
| Command R+ Refresh | 78.8 | 14.4 | 25.8 | 25.6 | 54.4 | 2.5 | 43.7 | 77.0 |
| Llama 3.3 70B Instruct | 86.0 / 81.0 | **32.9** | 46.9 / 41.9 | 47.8 | 75.5 | 3.2 | 46.2 | 85.6 |
| Mistral Large 2 | 84.7 | 26.7 | 44.7 | 54.0 | **82.9** | 10.8 | 46.8 | 88.0 |
| Qwen 2.5 72B Instruct | 88.6 | 26.3 | 45.8 / 43.6 | 48.3 | 78.5 | 6.3 | 55.7 | 83.2 |
| Llama 3.1 405B Instruct | 88.6 / 87.0 | 29.3 | 46.2 | 52.7 | 76.7 | 3.2 | 59.5 | 90.4 |
| DeepSeek V3 | **90.0** | **33.5** | 50.0 / 48.6 | 61.5 | **83.5** | 15.2 | 63.3 | 92.2 |

Table 12: **Code Understanding Benchmarks** across Python, Multi-language, and COBOL groups reporting 1-shot `pass@1` and RepoQA reporting match accuracy. HE is HumanEval. All results are internal reproductions using an identical prompt except where '/' indicates external value first and internal reproduction second. Best score ±1% is bolded. *For BigCodeBench, we use 3 tool-use execution feedback tests.

| | SWE-Bench Verified | Aider Polyglot |
|---|---|---|
| Command A | 26.8 | 14.7 |
| Command A Expert | 23.4 | 8.9 |
| Command R7B | 3.6 | 2.7 |
| Command R Refresh | 11.6 | 1.8 |
| Command R+ Refresh | 17.0 | 2.2 |
| Llama 3.3 70B Instruct | 29.4 | 8.4 |
| Mistral Large 2 | 30.0 | 16.0 |
| Qwen 2.5 72B Instruct | 33.0 | 8.0 |
| Llama 3.1 405B Instruct | 33.4 | 13.8 |
| DeepSeek V3 | 42.0 / **45.8** | 49.6 / **51.6** |

Table 13: **Code Editing Benchmarks**. All results are internal reproductions using an identical prompt except where '/' indicates externally reported value first and internal reproduction second.

**Code Understanding** evaluates code generation across multiple languages. For Python generation, we report on MBPP+ (Austin et al., 2021; Liu et al., 2024c), LiveCodeBench (Jain et al., 2024, Version 5 10/24-2/25), BigCodeBench (Zhuo et al., 2024, Instruct), and RepoQA (Liu et al., 2024b, 32K context length, threshold 0.8). For multi-language generation, we report HumanEval (Chen et al., 2021; Muennighoff et al., 2023) scores in Python, C++, Java, Javascript, Go, and Rust. We also extend our earlier uncontaminated Python benchmark, Less Basic Python Problems (Matton et al., 2024, LBPP), with parallel versions in C++, Java, Javascript, Go and Rust for uncontaminated generation evaluation across enterprise-critical programming languages.[5]

To assist in future advancements in COBOL understanding, we also develop a parallel version of HumanEval in COBOL (i.e., `HumanEval-COBOL`). We evaluate direct generation of COBOL, and translation of COBOL to Python similar to Muennighoff et al. (2023). The translation setting tests model capability to update legacy codebases into a modern language.

[5]We will release this dataset in an update to `huggingface.co/datasets/CohereForAI/lbpp`

| | Spider | | Bird | Internal | | | | | |
|---|---|---|---|---|---|---|---|---|---|
| | Dev | Test | Dev | Avg. | SQLite | PostgreSQL | MySQL | PL/SQL | T-SQL |
| Command A | 79.5 | 80.2 | **59.5** | 55.3 | 48.7 | 58.0 | 56.0 | **58.7** | 55.3 |
| Command A Expert | **85.5** | 85.4 | **58.5** | 56.1 | 49.3 | 60.0 | 55.3 | **58.0** | 58.0 |
| Command R7B | 78.1 | 77.6 | 42.2 | 34.4 | 27.3 | 36.0 | 34.7 | 38.7 | 35.3 |
| Command R Refresh | 76.5 | 78.1 | 47.3 | 42.8 | 36.7 | 48.0 | 42.7 | 43.3 | 43.3 |
| Command R+ Refresh | 82.0 | 81.7 | 52.7 | 44.4 | 40.7 | 47.3 | 40.0 | 52.0 | 42.0 |
| Llama 3.3 70B Instruct | 81.1 | 84.8 | 58.0 | 45.9 | 41.3 | 48.0 | 43.3 | 50.0 | 46.7 |
| Mistral Large 2 | 78.8 | 76.3 | 50.0 | 53.3 | 54.0 | 54.7 | 50.7 | 53.3 | 54.0 |
| Qwen 2.5 72B Instruct | 83.5 | 83.8 | 50.1 | 53.7 | 52.7 | 54.7 | 56.7 | 49.3 | 55.3 |
| Llama 3.1 405B Instruct | 83.0 | **86.7** | **59.4** | 49.2 | 54.0 | 58.7 | 50.7 | 34.0 | 48.7 |
| DeepSeek V3 | 81.7 | 81.7 | 53.1 | **60.8** | **56.7** | **66.0** | **60.7** | **58.7** | **62.0** |

Table 14: **SQL Generation Benchmarks** reporting execution accuracy against gold databases. All results are internal reproductions using an identical prompt. Avg. is the sample-weighted average across internal multi-dialect evaluation datasets. Best score ±1% is bolded.

**Code Understanding** metrics are outlined in Table 12. Sources of externally reported values are in Appendix B.3. Command A provides strong Python and multi-language performance compared to similar and larger models. Table 26 details the complete performance for HumanEval and LBPP in all languages, highlighting competitive accuracy in many business-critical programming languages. In the hardest benchmarks, LiveCodeBench and BigCodeBench, Command A surpasses many competitors and can be further improved with agentic tool-use discussed below. Command A also leads in RepoQA performance compared to all competitors. Finally, Command A offers state-of-the-art capabilities in COBOL for both direct generation, via HumanEval-COBOL, and translation from HumanEval-COBOL to HumanEval-Python. These strengths highlight that Command A offers accurate code understanding in the complex environment of navigating legacy enterprise codebases.

We also investigate the performance of Command A as a **code agent** using **multi-hop tool use** similar to the setup for RAG in Section 4.2. Command A can now access a code execution tool and receives feedback on code generation via execution results from gold-standard unit tests similar to Gehring et al. (2025). We evaluate 3 datasets in this regime: LiveCodeBench, BigCodeBench, and LBPP(all languages). In LiveCodeBench, we use the public unit tests for execution feedback and private tests for final evaluation. For BigCodeBench and LBPP, we simulate the unit-test split by using 3 unit tests for execution feedback and all remaining tests for final evaluation.[6] Table 12 shows how using Command A as an agent easily surpasses direct code generation across all datasets—achieving a `pass@1` gain over Command A of +5.9% for LiveCodeBench, +14.3% for BigCodeBench, and +12.3% for LBPP across all languages. Notably, Command A achieves 71.4% in LBPP-Python surpassing all competitors by 4.3% and surpasses all other models in the BigCodeBench leaderboard at the time of publication.[7]

**Code Editing** evaluates the model capability to generate precise code line-level changes to edit and update a codebase. We evaluate our models on the SWEBench Verified Patch Generation task in Python (Jimenez et al., 2024), and the Aider Polyglot benchmark[8] for multi-language code editing in Python, C++, Java, Javascript, and Rust. Table 13 demonstrates Command A is competitively capable in repository-level understanding and solving pull-requests or building code fixes via patch generation. We note that these results are from post-hoc investigations into code-editing behaviour in our model as we did not target these functions in developing Command A. Similar to our investigation into a code agents described above, we share these results as early signposts for future objectives of code expert development.

[6]As the prompt design for LBPP includes 3 unit-tests, this setup does not leak any further testing requirements to the model.

[7]The current best model is `GPT-4o-2024-05-13` with 51.1 `pass@1` `https://bigcode-bench.github.io/`

[8]`aider.chat/2024/12/21/polyglot.html`

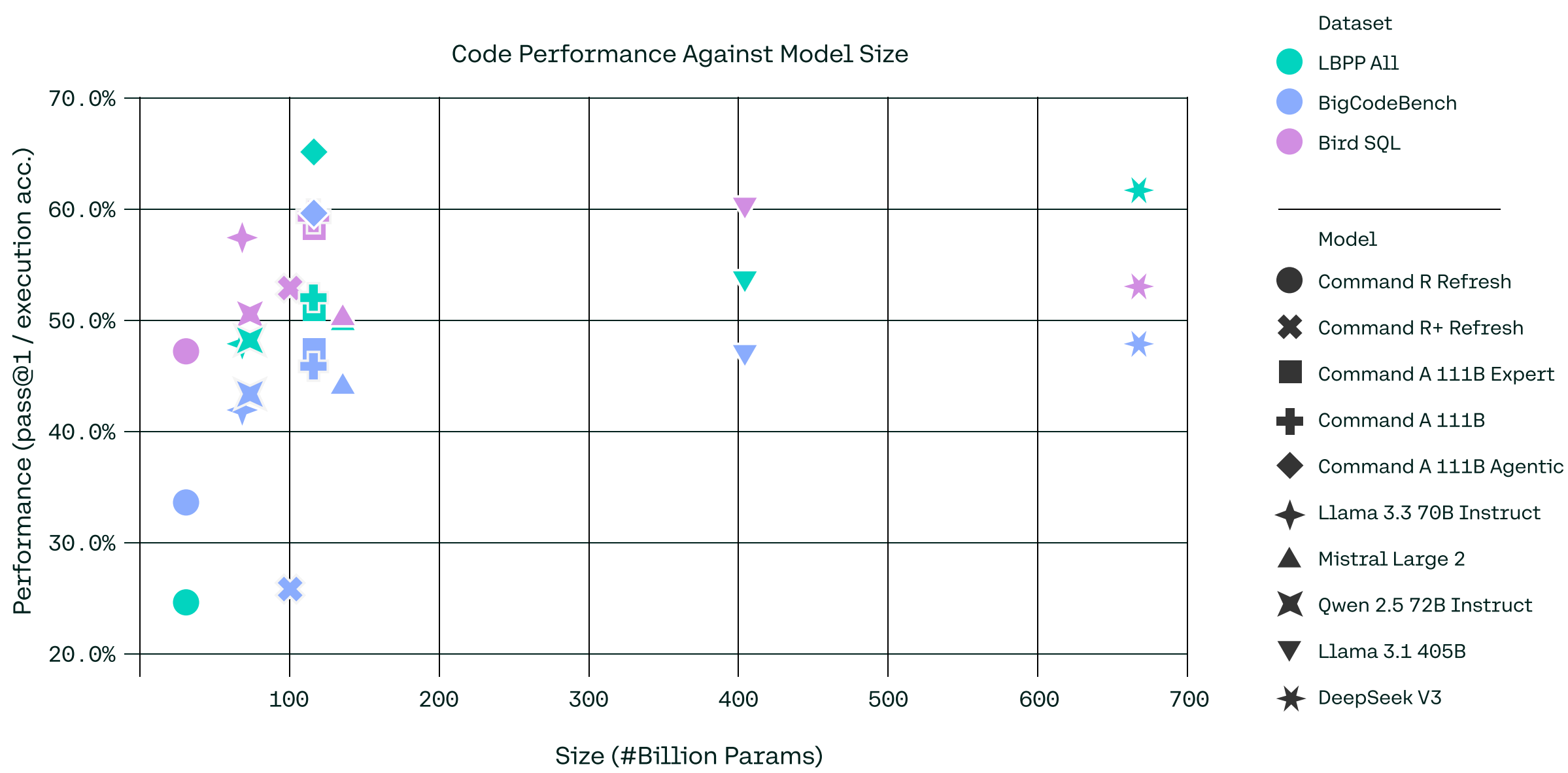

Figure 6: **Code Performance against Model Size.** Command A provides state-of-the-art performance compared to models of similar size, and often significantly larger models. Command A Agent improves even further to set a new standard for performance at 111B size with tool-use in code.

**SQL Generation** evaluates model capability in understanding user requests using a partially observed database context. Understanding SQL and reasoning with databases is critical for Command A to succeed as an enterprise model. We evaluate SQLite performance using Spider (Yu et al., 2018, Dev & Test) and the more recent Bird SQL benchmark (Li et al., 2023a, Dev). To ensure Command A can accurately generate SQL in an enterprise database context, we also report results for an internal benchmark in SQLite, PostgreSQL, MySQL, Oracle PL/SQL, and Microsoft T-SQL. Performance on these dialects better reflects real usage of SQL to access commercial database systems.

Table 14 demonstrates that Command A offers state-of-the-art performance across multiple datasets. Command A leads in both Spider Dev, and Bird to provide accurate SQL generation to solve challenging queries in even "dirty" database contexts. Across models of similar size, Command A also demonstrates the strongest average performance across 5 enterprise-critical SQL dialects in our internal benchmark. This further punctuates the capability of Command A in both academic and enterprise scenarios for SQL.

We highlight the performance benefit of Command A relative to size in Figure 6. Across 3 datasets, Command A and Command A Code Expert provide best-in-class performance, often surpassing similar and larger models. Command A offers a unique trade-off for enterprise capability in accurate code and SQL generation. Using Command A as an agent for code further enhances the model for state-of-the-art capabilities across challenging benchmarks.

## 4.5 Math and Reasoning

We evaluate the reasoning capability of our model on key mathematical reasoning benchmarks, and compare this to publicly-reported metrics (where available) in Table 15. We find that Command A performs especially well on mathematical benchmarks, and that merging models preserves reasoning performance (compared to reasoning-expert models) within a few percentage points across most benchmarks (§4.9).

| | MATH (all) | AIME (2024) | GPQA (Diamond) |
|---|---|---|---|
| Command A | 80.0 | 23.3 | 50.8 |
| GPT-4o | 68.5 | 9.3 | 46.0 |
| Llama 3.3 70B | 77.0 | 20.0* | 50.5 |
| Llama 3.3 405B | 73.9 | 20.0* | 49.0 |
| Mistral Large 2 | 71.3* | 11.0 | 48.6 |

Table 15: Reasoning performance of Command A compared to similarly-sized models. Benchmarks are MATH (Hendrycks et al., 2021), the 2024 AIME mathematics competition, and GPQA Diamond (Rein et al., 2023). Results for external models are taken from officially-reported sources, unless indicated with an asterisk (*), which denotes internal evaluation since official public results were not available.

In our qualitative assessments, we also find that reasoning-expert models provide generalised gains in coding and structured data manipulation tasks, and that these are additive in the final Command A model.

## 4.6 Safety

Our safety evaluation methodology combines human and automated assessments. Due to speed and cost considerations, we mainly rely on automated evaluations. We use human annotations as a baseline to ensure our automated evaluations align with human judgment. These are triply annotated by an internal team of specialist safety annotators. To further strengthen the reliability of our automated evaluation, we assess the suitability of evaluators based on their robustness to artifacts (Chen & Goldfarb-Tarrant, 2025).

We measure both **absolute** and **relative** safety. Absolute safety evaluation tests models with potentially eliciting prompts from the categories of our core safety behaviour (§3.3.7), and then computes the rate of unsafe content in the model output, using an LLM-as-a-judge setup. The absolute safety aggregate score is the average of each categorical rate, where each category is weighted equally. Relative safety evaluation uses the same prompts, but considers how the safety of each response compares to the safety of another model's response for the same prompt. If both responses are equally safe, the higher quality response is chosen as the winner. Relative safety is more challenging, so we rely on a jury of LLM evaluators (Verga et al., 2024), which achieves human agreement scores of 77.7% and Cohen's Kappa of 0.55 in relative safety evaluations.

We also measure **over-refusal** rate; how frequently models refuse to answer a prompt that should be answered. These prompts fall into two categories: word sense disambiguation and requests for information about safety topics. We use an LLM-as-a-judge setup, as we find refusal classification a much easier task than safety, with very high accuracy and human agreement

### 4.6.1 Enterprise Safety

In enterprise contexts, safety priorities and risk assessments differ significantly from those in default contexts. We consider two elements that are of strong concern for Enterprise usage: **Controllability**, the ability of the model to be customised for different safety needs, and **Demographic Fairness**, the robustness of the model to demographic perturbations in tasks involving real human data.

#### *4.6.1.1 Controllability*

In Enterprise Safety, the notion of safety itself is context-dependent. Some core safety behaviour is consistent across all contexts (§3.3.7.1), but much of it varies between different deployments. The boundaries of content that an LLM should generate when used as an LLM-editor for a journalist are very different than the content boundaries of a customer service chatbot. Therefore, we evaluate the model's ability to accurately condition on different safety instructions, under our two safety modes: contextual and strict (§3.3.7.1). For each mode we compose two evaluation sets: one that should always be answered (over-refusal evaluation) and one that should always be refused (safety mode control), which allows us to optimise the trade-off between these two scenarios.[9] Safety mode accuracy is the mean of these sets for a given mode.

[9]We note that the over-refusal evaluation set was created by red-teaming Command R+ Refresh.

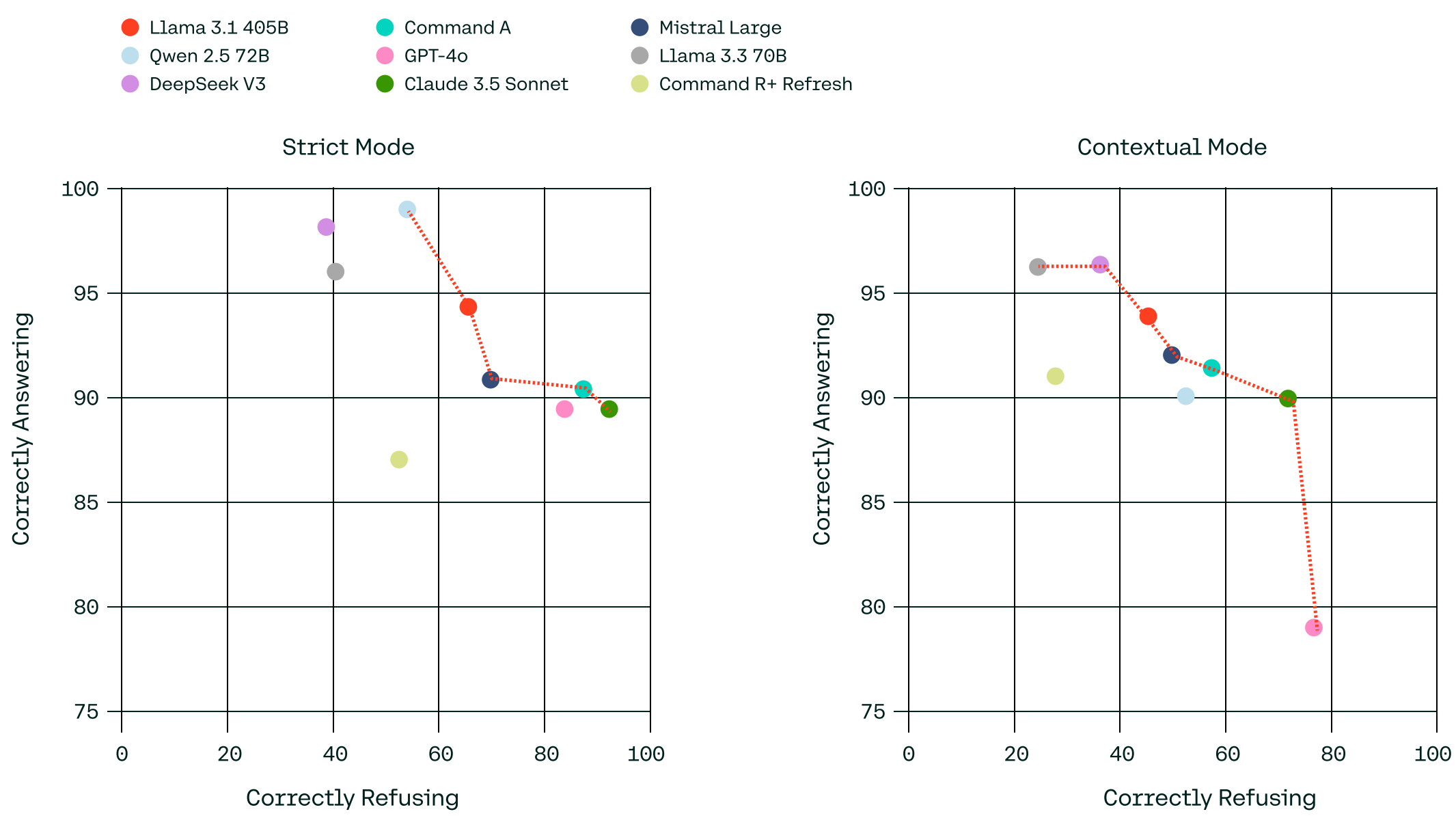

Figure 7: The Pareto frontier between correctly answering and refusing for our enterprise safety modes.

Figure 7 shows that the Command A model is on the Pareto frontier between answering and refusing for both safety modes. Results for competitor models can be found in appendix Table 27. Each competitor targets different markets and behaviours, so we consider different modes to have effectively different competitors. In contextual mode, the relevant competitors are Mistral Large 2, Qwen 2.5 72B Instruct and Llama 3.3 70B Instruct, while in strict mode the relevant competitors are GPT-4o and Claude 3.5 Sonnet.

#### *4.6.1.2 Demographic Fairness*

LLMs are used in various hiring software systems in the market, and we evaluate demographic fairness in this context. The model is tasked with summarising the suitability of resumes with respect to a given job description. We follow Seshadri & Goldfarb-Tarrant (2025) for both our method and our metric. We permute the demographics of the resume and measure meaningful differences in generated summaries for candidates when their race or gender has changed. A perfect model would have no meaningful differences, i.e. would be invariant to the perturbation. The bias metric is defined as the proportion of measurements (including reading ease, subjectivity and regard, as outlined in Seshadri & Goldfarb-Tarrant (2025) for which the null invariance hypothesis is rejected when comparing the original and perturbed summaries. To account for variability in generations (Chen & Goldfarb-Tarrant (2025) observed this even at temperature 0), we generate responses using each model five times per sample and plot the distribution of bias rates across all runs. The results for gender and race are shown in Figure 8. We report with both Bonferroni (bonf) and Benjamini-Hochberg (bh) corrections to account for the multiple measurements on the same summaries and to allow the reader to select whichever correction is more applicable – bonf to minimise false positives (finding a demographic fairness issue when there is none), and bh to minimise false negatives.

We note two broad patterns across all models: models tend towards much stronger racial bias than gender bias, and smaller models tend to have greater bias than larger models. In particular, the Command A models show impressive robustness to demographic perturbations. Command A is entirely robust to gender perturbations and very resilient to race ones (only 1% failures). Command R7B similarly is entirely robust to gender in this evaluation, and competitive for a small model at robustness to race, with around 4% failures.

We don't observe significant gender bias for large models in this domain in our testing setup. Command A, Llama 3.3 70B Instruct, and Mistral Large 2 all exhibit minimal racial bias, each failing a median of 1% of invariance tests, while Claude 3.5 Sonnet has the lowest, at 0%. Small models are significantly less robust.

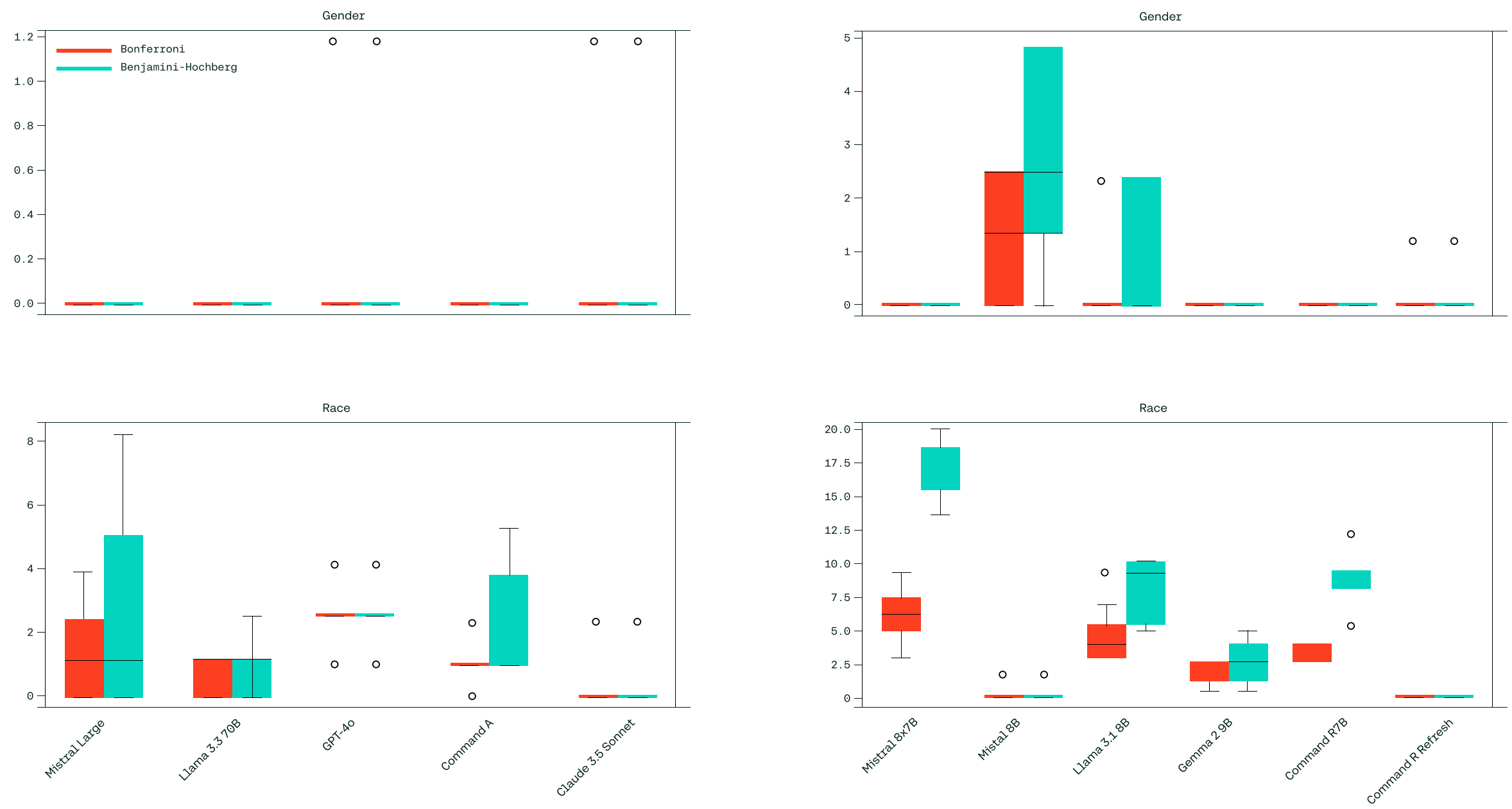

Figure 8: Boxplots of gender and racial bias rates in model-generated resume summaries for Command A (left) and Command R7B (right) compared to similarly sized models, respectively, using either Bonferroni or Benjamini-Hochberg correction. The Command A models show impressive robustness to demographic perturbations. Command A is robust to gender perturbations and very resilient to race ones (only 1% failures). Command R7B similarly is robust to gender in this evaluation, and competitive for a small model at robustness to race, with around 4% failures.

Most small models remain robust to gender, with the exception of Llama 3.1 8B Instruct and Ministral 8B, which fail 1-5% of invariance tests. Interestingly, Ministral 8B lacks robustness to gender, but is robust to race, whereas Mixtral lacks robustness to race, but is robust to gender.

Overall, our models offer excellent coverage of robustness across different demographic categories, for multiple sizes. We note that, though generation does contribute, total demographic fairness in a hiring pipeline is dominated by the retrieval stage (Seshadri & Goldfarb-Tarrant, 2025). Here we measure only the generation stage, but our embedding model for the retrieval stage is also the most robust to perturbations.

#### 4.6.2 Default Safety

In the default setting, we evaluate the safety of the model without a system preamble to simulate cases outside of Cohere's API or enterprise contexts.

Command A shows strong performance in various categories of unsafe content. As shown in Figure 9, Command A significantly outperforms all competitors in relative safety evaluations. Additionally, it attains an absolute safety score of 70.4%, ranking third among large models, closely following Claude 3.5 Sonnet and Qwen 2.5 72B Instruct (Table 16). It excels at avoiding violence and hate speech, with a 89.7% safe response rate, and performs well in areas such as not generating CSEA (87.5%) and not promoting misinformation (67.9%) (Figure 15). While Command R7B shows lower overall performance, it still maintains a notable presence in certain categories, such as avoiding violence and hate speech (76.3%) and not promoting CSEA (67.0%) (Figure 16). These results highlight the effectiveness of Command A in mitigating unsafe content generation even in situations where we cannot add system preamble guardrails.

Although the relative and absolute safety performance of Command A may initially seem contradictory, this occurs because the relative safety evaluation considers the intersection of safety and quality. Critically, in the event that both models provide a safe response, the relative safety evaluations then consider the winner to be the model that provides a higher quality response. Rather than simply refusing to answer, Command A engages meaningfully with queries that relate to potentially unsafe topics. Many other models, such as Claude 3.5 Sonnet, provide non-specific refusals.

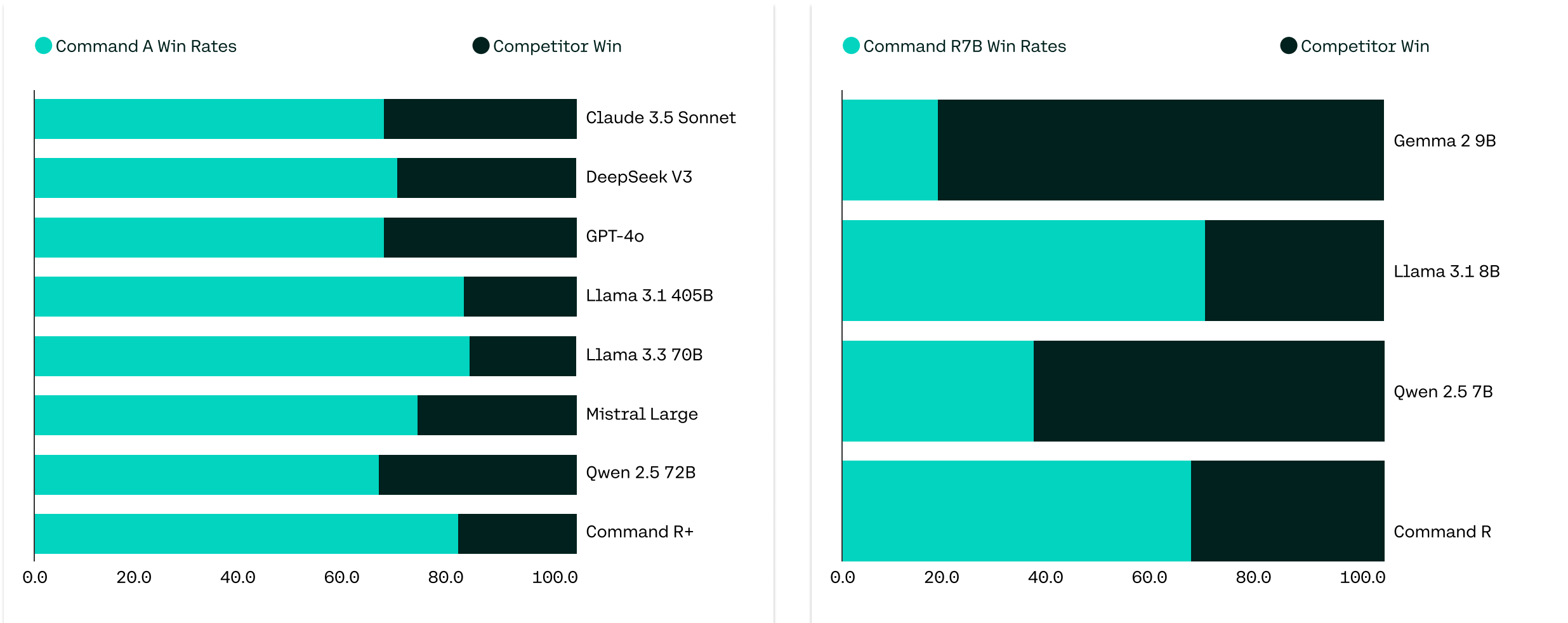

(a) Large Models

(b) Small Models

Figure 9: **Default relative safety performance.** Winner is assigned by a panel of LLM judges. When both responses are equally safe, the winner is chosen based on which response is higher quality.

| | Relative Safety(↑) | Absolute Safety(↑) | Misinfo(↑) | Self-harm(↑) | CSEA(↑) | Sexual Content(↑) | Violence & Hate(↑) |
|---|---|---|---|---|---|---|---|
| Command A | **49.5** | 70.4 | 67.9 | 61.2 | 87.5 | 63.1 | 89.7 |
| Claude 3.5 Sonnet | 26.4 | **80.0** | **76.9** | **90.4** | **98.5** | 94.4 | **93.1** |
| DeepSeek V3 | 23.7 | 49.7 | 50.0 | 37.0 | 74.1 | 34.8 | 74.0 |
| GPT-4o | 26.6 | 65.6 | **76.9** | 69.9 | 33.7 | **95.5** | 84.4 |
| Llama 3.1 405B | 15.9 | 41.8 | 42.3 | 28.8 | 63.0 | 34.3 | 62.2 |
| Llama 3.3 70B | 14.9 | 40.5 | 50.0 | 42.5 | 63.0 | 6.7 | 61.8 |
| Mistral Large 2 | 22.0 | 45.7 | 57.7 | 37.0 | 74.8 | 8.0 | 71.0 |
| Qwen 2.5 72B | 32.4 | 71.4 | 61.5 | 60.3 | 86.3 | 91.6 | 90.0 |
| Command R+ Refresh | 16.3 | 30.2 | 42.3 | 21.9 | 45.9 | 5.1 | 47.3 |
| Command R7B | 49.7 | 58.2 | 50.0 | 50.7 | 67.0 | 47.2 | 76.3 |
| Gemma 2 9B | **81.3** | **87.3** | **76.9** | **82.2** | **94.8** | 85.4 | **96.9** |
| Llama 3.1 8B | 35.9 | 63.6 | 57.7 | 58.9 | 60.7 | 66.3 | 74.4 |
| Qwen 2.5 7B | 59.2 | 71.5 | 65.4 | 50.7 | 71.9 | 87.6 | 82.1 |
| Command R Refresh | 30.4 | 31.3 | 38.5 | 26.0 | 37.8 | 3.9 | 50.4 |

Table 16: **Default safety performance** of Command A and Command R7B compared to similarly sized models across various categories of unsafe content. Relative safety is the winrate vs. Command A. Absolute safety score is computed as an average of safe response rates for all categories. Large models are shown in the top half of the table, while small models are shown in the bottom half. The top performing model for each size category is bolded in each column. As indicated by the upwards-pointing arrows, higher winrates and higher safe response rates correspond to better performance for each competitor.

We also measure over-refusal rates for the default setting on the XSTest benchmark (Röttger et al., 2024). Command A shows refusal rates under 3%, which is considerably better than other closed-source models, namely Claude 3.5 Sonnet and GPT-4o; and marginally better than open-access models such as Llama 3.3 70B Instruct (see Table 17). As XSTest is saturated, we also report default over-refusal based on our internal test set from red-teaming our previous model.

| | XSTest Refusal (↓) | | Over-Refusal (↓) | |
|---|---|---|---|---|
| | Full | Partial | XSTest | Internal |
| Command A | 1.1 | 0.0 | 1.1 | 7.1 |
| Claude 3.5 Sonnet | 3.6 | 0.0 | 3.6 | 4.1 |
| DeepSeek V3 | 0.8 | 0.0 | 0.8 | 1.2 |
| GPT-4o | 5.6 | 0.0 | 5.6 | 7.1 |
| Llama 3.1 405B | 1.2 | 0.0 | 1.2 | 2.4 |
| Llama 3.3 70B | 1.2 | 0.0 | 1.2 | 3.5 |
| Mistral Large 2 | 0.8 | 0.0 | 0.8 | 2.4 |
| Qwen 2.5 72B | 0.4 | 0.0 | 0.4 | 1.2 |
| Command R+ Refresh | 3.6 | 1.2 | 4.8 | 5.3 |
| Command R7B | 4.0 | 0.8 | 4.8 | 11.8 |
| Gemma 2 9B | 2.8 | 4.0 | 6.8 | 24.1 |
| Llama 3.1 8B | 6.4 | 0.0 | 6.4 | 8.8 |
| Qwen 2.5 7B | 0.8 | 0.0 | 0.8 | 9.4 |
| Command R Refresh | 3.2 | 0.0 | 3.2 | 2.4 |

Table 17: **Over-refusal rate** of Command A and Command R7B, based on the XSTest benchmark, which distinguishes between full refusal and partial refusal. We sum both to obtain the over-refusal rate.

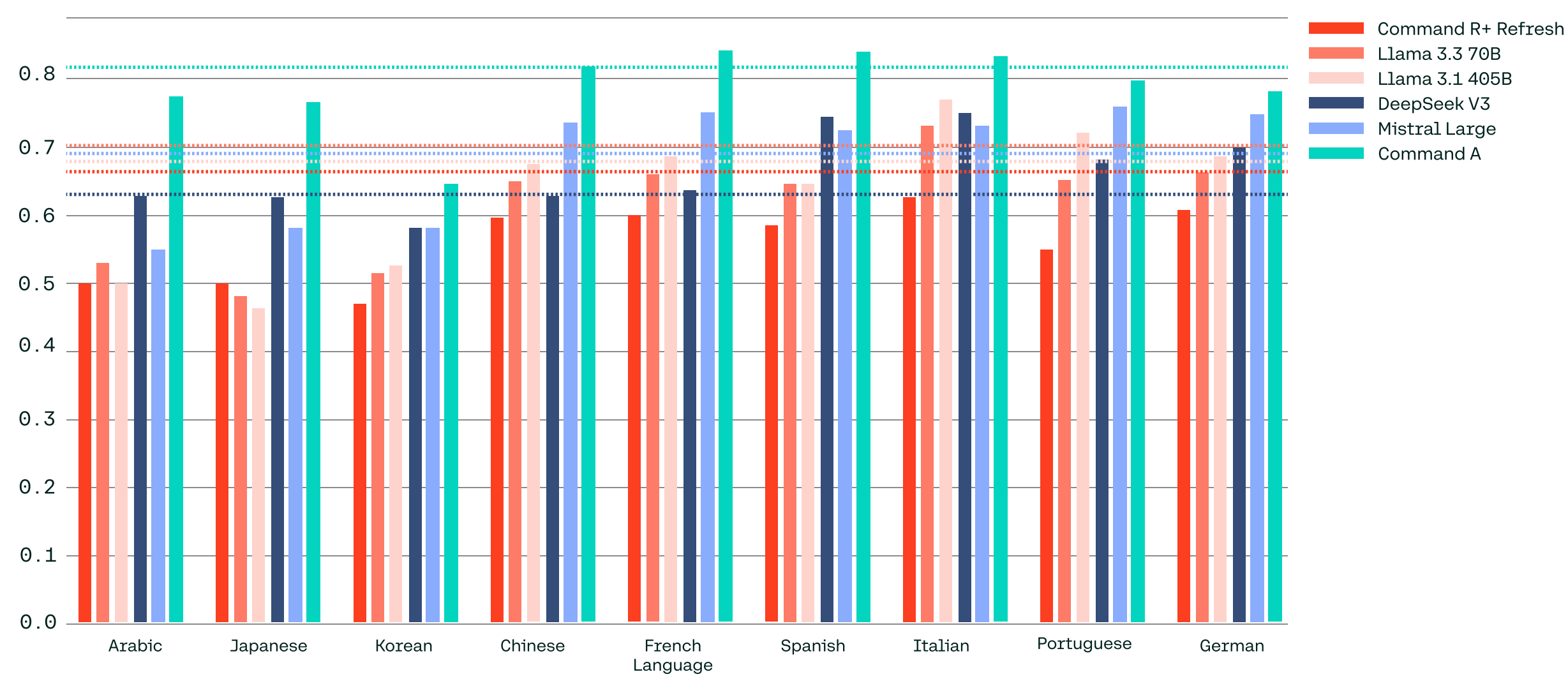

Figure 10: Safety scores (rate of safe responses averaged over Misinformation and Violence and Hate) across 9 languages. The dashed lines show the English score.

#### 4.6.3 Multilingual Safety

We evaluate safety on nine priority languages (results in Figure 10). The safety score is the rate of safe responses measured on the same set of prompts across all languages, thus allowing for direct comparisons. This set is the Misinformation and Violence and Hate categories from English, translated automatically, then corrected and naturalised by multilingual annotators. We use LLM-as-a-judge evaluation. For some languages, completions are translated into English before being evaluated, and some are retained in the original language, based on which showed the best performance on a development set.

We also evaluate over-refusal via a prompt set collected through red teaming with the multilingual annotators. The English over-refusal set was translated into each language, then refined and augmented with more natural prompts. Command A is on the Pareto frontier between safety and over-refusal in 5 languages (Japanese, Chinese, German, Korean and Arabic), whilst remaining competitive in all other measured languages.

## 4.7 Enterprise-specific Benchmarks

Our enterprise evaluation process focuses on testing model capabilities in generative and retrieval-augmented use cases that mirror typical enterprise applications.

**Generative use cases.** We use rule-based checks and LLM-as-a-judge evaluations due to the complexity and nuances of the differing enterprise use cases. We show an example prompt for reference in Figure 11.

Example Prompt

Create a job ad for the role of Social Media Manager at the BuzzMedia, located in New York, NY. The ad should be creative and engaging to attract the best talent. It should include a catchy and creative title, a brief role overview, and a list of at least 5 employee benefits offered by the company. The overview should be at least 50 words.
Format the response as a JSON object in the following format.

```
{
  "title": <title>,
  "role_overview": <role_overview>,
  "employee_benefits": [
    <Benefit 1>,
    <Benefit 2>,
    <Benefit 3>,
    <Benefit 4>,
    <Benefit 5>
  ]
}
```

Figure 11: Example prompt for an enterprise generative use case.

We break down success criteria into rule-based and LLM-based checks:

- **Rule-Based Checks.** Checks for attributes like word count, valid JSON output, count of expected items, and similar.
- **LLM-Based Checks.** We use a panel of judges, similar to the approach described in Verga et al. (2024), to evaluate more nuanced criteria like tone and natural language quality.

The final per-example score is an average of the rule- and LLM-based scores. Our generative enterprise benchmark consists of 22 tasks covering use cases including chat and meeting summarisation, information extraction, and FAQ generation. A summary of the results across all tasks can be found in Table 18. Command A achieves the highest pass rate at 94.2% across all generative use cases. Command R7B scores 71.9%, which is the highest performance among similarly-sized models.

**RAG use cases.** For enterprise RAG evaluation tasks, we assess question-answering use cases involving technical documentation and workplace policies, including internal rules and company benefits. These tasks often involve user queries on long document snippets that can exceed 10,000 tokens. Some questions require synthesizing information from multiple snippets, while others cannot be directly answered using the given documents. Our evaluation set includes ground-truth answers annotated by humans.

To assess the performance of our models, we use two key evaluation metrics:

- **Correctness**: Measured using Llama Index Correctness, this evaluates the validity of a model's response. It ensures that generated answers align with the provided context and are factually correct.
- **Answerability Accuracy**: Assessed by LLM judges, this metric measures the model's ability to discern between answerable and unanswerable questions.

| Model | Enterprise Pass Rate (%) |
|---|---|
| Command A | 94.2 |
| Command R+ Refresh | 87.4 |
| Claude 3.5 Sonnet v2 | 84.2 |
| DeepSeek V3 | 81.3 |
| GPT-4o | 79.1 |
| Llama 3.3 70B Instruct | 65.2 |
| Llama 3.1 405B Instruct | 77.8 |
| Command R7B | 71.9 |
| Gemma 2 9B | 65.7 |
| Llama 3.1 8B | 60.4 |

Table 18: Enterprise generative evaluation performance across all 22 tasks.

| Model | Workplace Policies QA | Technical QA | Avg. |
|---|---|---|---|
| Command A | 4.59 | **4.86** | **4.73** |
| Command R+ Refresh | 4.08 | 4.42 | 4.25 |
| Claude 3.5 Sonnet v2 | **4.63** | 4.81 | 4.72 |
| DeepSeek V3 | 4.52 | 4.64 | 4.58 |
| GPT-4o | 4.50 | 4.81 | 4.66 |
| Llama 3.3 70B Instruct | 4.45 | 4.78 | 4.61 |
| Llama 3.1 405B Instruct | 4.41 | 4.62 | 4.52 |
| Command R7B | **4.16** | 4.48 | **4.32** |
| Gemma 2 9B | 3.23 | **4.66** | 3.95 |
| Llama 3.1 8B | 3.99 | 4.37 | 4.18 |

Table 19: Enterprise RAG evaluation performances of Command A models. The table shows the LLama Index Correctness metric which ranges from 1-5. Best performances are marked in bold.

| Model | Answerable Acc. (%) | Unanswerable Acc. (%) |
|---|---|---|
| Command A | **96** | 91 |
| Command R+ Refresh | 78 | **95** |
| Claude 3.5 Sonnet v2 | 89 | 92 |
| DeepSeek V3 | 93 | 86 |
| GPT-4o | 94 | 88 |
| Llama 3.3 70B Instruct | 95 | 87 |
| Llama 3.1 405B Instruct | 94 | 90 |
| Command R7B | 86 | 76 |
| Gemma 2 9B | **90** | **90** |
| Llama 3.1 8B | 88 | **91** |

Table 20: Enterprise RAG Answerable and Unanswerable Accuracy.

These metrics ensure that the model generates accurate responses while demonstrating robust judgement in handling questions beyond the scope of the provided context. Table 19 shows the LLama Index Correctness performance on RAG use cases and Table 20 shows the Answerable and Unanswerable Accuracy across tasks. The accuracy across answerable and unanswerable questions is typically a trade-off as we do not want the model to over-refuse when the question is actually answerable, given the context.

Command A has the best Llama Index Correctness Average at 4.73, Answerable Accuracy of 96% and an Unanswerable Accuracy of 91%, indicating that it responds when intended, while keeping hallucination rates low for unanswerable questions. Command R7B has the best Llama Index Correctness Average score at 4.32 compared to models of similar size, an Answerable Accuracy of 86% and an Unanswerable Accuracy of 76%.

## 4.8 Long-Context Benchmarks

To assess long-context understanding capability, we employ two extensive long-context benchmark datasets: RULER (Hsieh et al., 2024) and LongBench-V2 (Bai et al., 2025). RULER comprises 13 distinct tasks, including retrieval, question-answering, multi-hop tracing and aggregation tasks. It is designed to evaluate a model's ability to retrieve and reason over longer context inputs. The evaluation is conducted on sequences of up to 256k tokens. LongBench-V2 includes a diverse set of question-answering tasks spanning various levels of difficulty and multiple context types, such as single- and multi-document contexts, multi-turn dialogues, code repositories, and long structured data.

| | **4k** | **8k** | **16k** | **32k** | **64k** | **128k** | **256k** | **Avg** (≤128k) | **wAvg.** (inc) | **wAvg.** (dec) |
|---|---|---|---|---|---|---|---|---|---|---|
| Command A | 97.2 | 96.9 | 96.7 | 95.9 | 93.3 | 90.0 | 84.6 | **95.0** | **93.9** | **96.1** |
| Mistral Large 2 | 96.4 | 96.3 | 95.3 | 94.0 | 85.9 | 48.1 | - | 86.0 | 79.5 | 92.5 |
| Llama 3.1 70B | 96.5 | 95.8 | 95.4 | 94.8 | 88.4 | 66.6 | - | 89.6 | 85.5 | 93.7 |
| Command R+ Refresh | 96.0 | 95.1 | 94.0 | 92.4 | 85.4 | 64.6 | - | 87.9 | 83.4 | 92.4 |
| GPT-4o(11-20) | 97.0 | 92.1 | 89.0 | 88.8 | 88.4 | - | - | - | - | - |
| Claude 3.5 Sonnet (10-22) | 96.5 | 96.0 | 95.7 | 95.0 | 95.2 | 93.8 | - | **95.4** | **95.0** | **95.8** |
| Gemini-1.5-Pro (002) | 96.2 | 96.0 | 96.0 | 95.8 | 93.8 | 91.7 | 91.6 | **94.9** | **94.2** | **95.6** |
| Gemini-2.0-Flash (exp) | 96.0 | 96.0 | 95.1 | 95.7 | 93.7 | 86.0 | 79.7 | 93.8 | 92.4 | 95.1 |

Table 21: Results on the RULER long context benchmark. Not all models support contexts up to 256k.

| | **Overall** | **Short** | **Medium** | **Long** | **Easy** | **Hard** |
|---|---|---|---|---|---|---|
| Command A | **43.4** | 44.1 | 47.4 | 34.3 | 45.3 | 42.3 |
| Mistral Large 2 | 34.4 | 41.7 | 30.7 | 29.6 | 38.0 | 32.2 |
| Llama 3.1 70B | 31.8 | 37.2 | 28.8 | 28.7 | 35.4 | 29.6 |
| Llama 3.1 405B | 37.8 | 39.7 | 39.0 | 32.1 | 39.0 | 37.0 |
| Command R+ Refresh | 27.8 | 36.7 | 23.7 | 21.3 | 30.2 | 26.4 |
| GPT-4o(11-20) | **46.0** | 42.9 | 52.1 | 35.3 | 52.1 | 42.2 |
| Claude 3.5 Sonnet (10-22) | 40.7 | 44.9 | 41.3 | 31.6 | 45.8 | 38.6 |

Table 22: LongBench-V2 results for Command A.

Tables 21 and 22 highlight Command A's exceptional long-context capabilities. Our hybrid architecture enables this level of performance while requiring significantly less KV cache memory compared to models with a full attention architecture. For instance, at an 8k sequence length, Command A requires only 75% of the KV cache memory used by Llama 3.3 70B Instruct, 23.8% of that used by Llama 3.1 405B Instruct, and 45.5% of that used by Mistral Large. At a 128k sequence length, these decrease to 32.9%, 10.4%, and 19.9%, respectively. This reduction in KV cache memory usage can significantly decrease latency and memory consumption while enhancing throughput during inference, particularly for longer contexts.

## 4.9 Merging

### 4.9.1 Expert performance is largely preserved

We find that model merging is an effective method for combining capabilities from a set of expert models into a single model, and that linear merging is sufficient to preserve expert performance with only a 1.8% average drop. Figure 12a shows the distribution of changes in score for the metrics tracked during merging. We find that the overwhelming majority of metrics are preserved to within 2.5% of the best expert score, with some metrics actually improving after merging. Figure 12b shows the average degree of metric preservation between expert and merge, grouped by domain. RAG and general performance are generally best preserved, with code performance the least well preserved during merging.

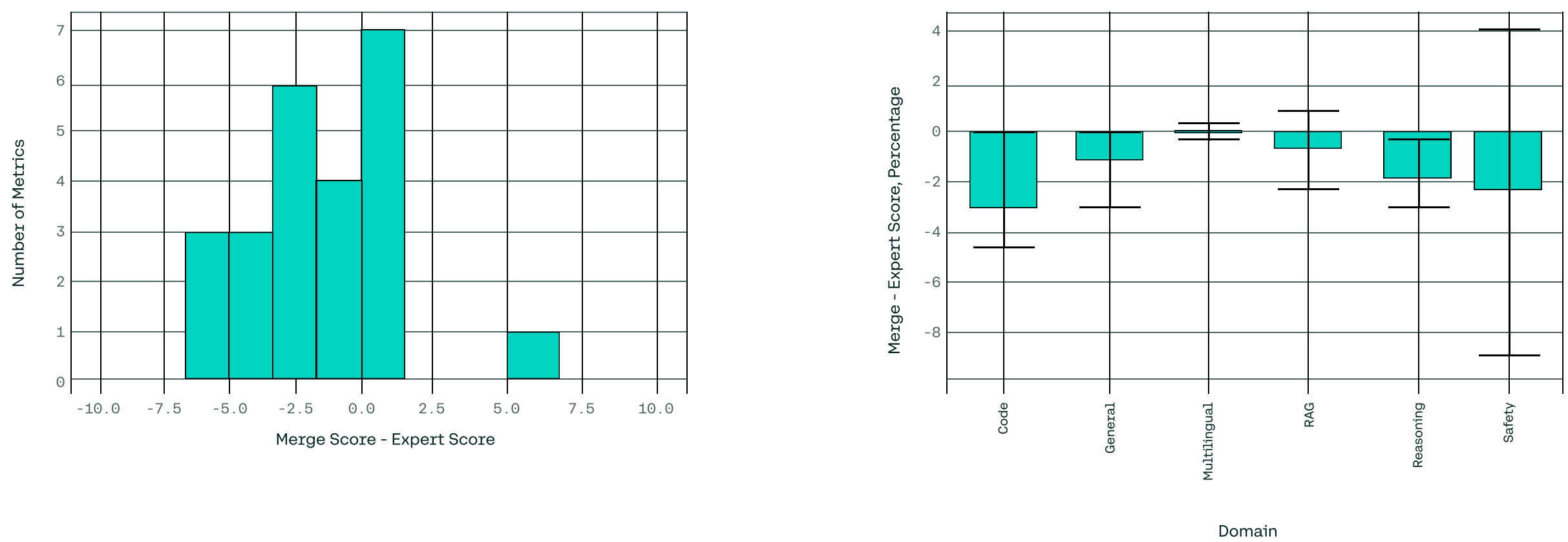

(a) Distribution of metric values for the merged model, as a percentage of the score for the best input expert. The overwhelming majority of metrics tracked are preserved to within 2.5% of the best expert score.

(b) Average degree of metric preservation between expert and merge, by domain. Error bars denote standard deviation. RAG and general performance are best preserved, with code the least well preserved. Note that the higher RAG standard deviation is a property of RAG metrics (rather than the merge) as TauBench has very high variance across runs.

Figure 12

We additionally find that, at the 111B scale, model merging is very stable. Large changes to expert weights result in relatively small (2-3%) changes in domain performance, with very few candidates displaying catastrophic failures at any capability.

### 4.9.2 Merging is a coarse operation

Model merging is a coarse operation, with no guarantee that a merged checkpoint will fall at a local optimum in the loss landscape. We find that applying a small number of SFT steps with low learning rate[10] on top of a merged checkpoint leads to significant improvements in model capability, with many metrics reaching or exceeding 100% of the expert performance. Intuitively, model merging appears to be effective at combining capabilities but may leave them 'misaligned' in encoding space; a small number of gradient-based updates gently adjusts these encodings and allows the capabilities to be fully surfaced.

## 4.10 Polishing

The polishing effort includes several phases to improve model style and overall alignment with human preferences. We use the same evaluation setting as the multilingual mArenaHard restricted to English (Dang et al., 2024). In Figure 13, we show Command A win rates against GPT-4o (1120) according to a pool of LLM judges at each phase of the CoPG/SRPO ping-pong. While this method allows us to achieve high win rates, we also observe that interleaving the two methods tends to increase overall training stability: a potential regression occurring at a particular phase is likely to be corrected by the next one.

In addition, polishing helps improve both the overall alignment with human preferences and recovery of any degraded scores during merging. For some of the metrics, it also improves the score over the expert models. Figure 14 provides the difference in metrics across many domains between the polished and the RL soup model, showing that some domains benefit significantly from polishing, with human preferences benefiting the most.

[10] Roughly 10-20x lower than was used for the main SFT training.

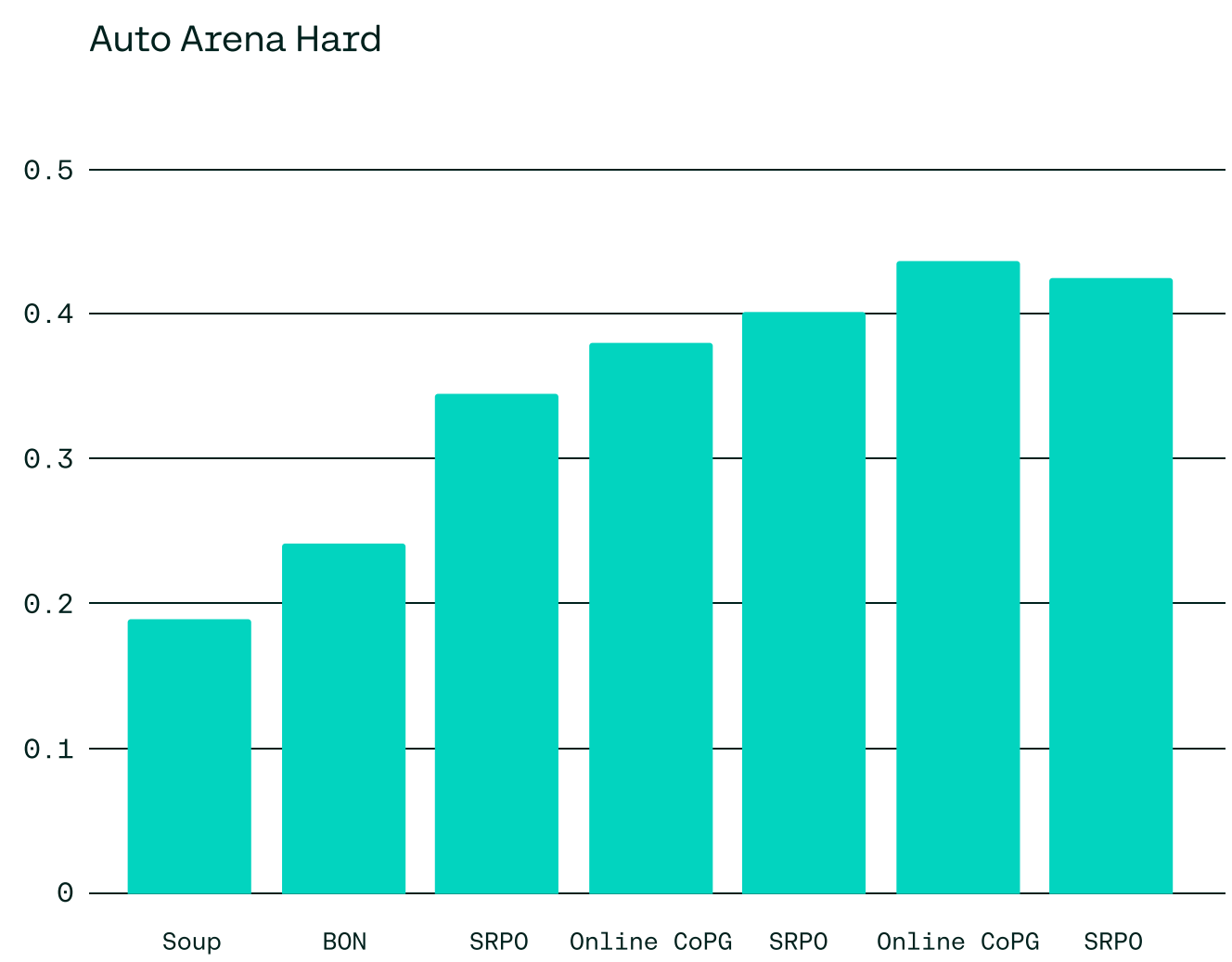

Figure 13: Command A win rates against GPT-4o as the polishing phase progresses.

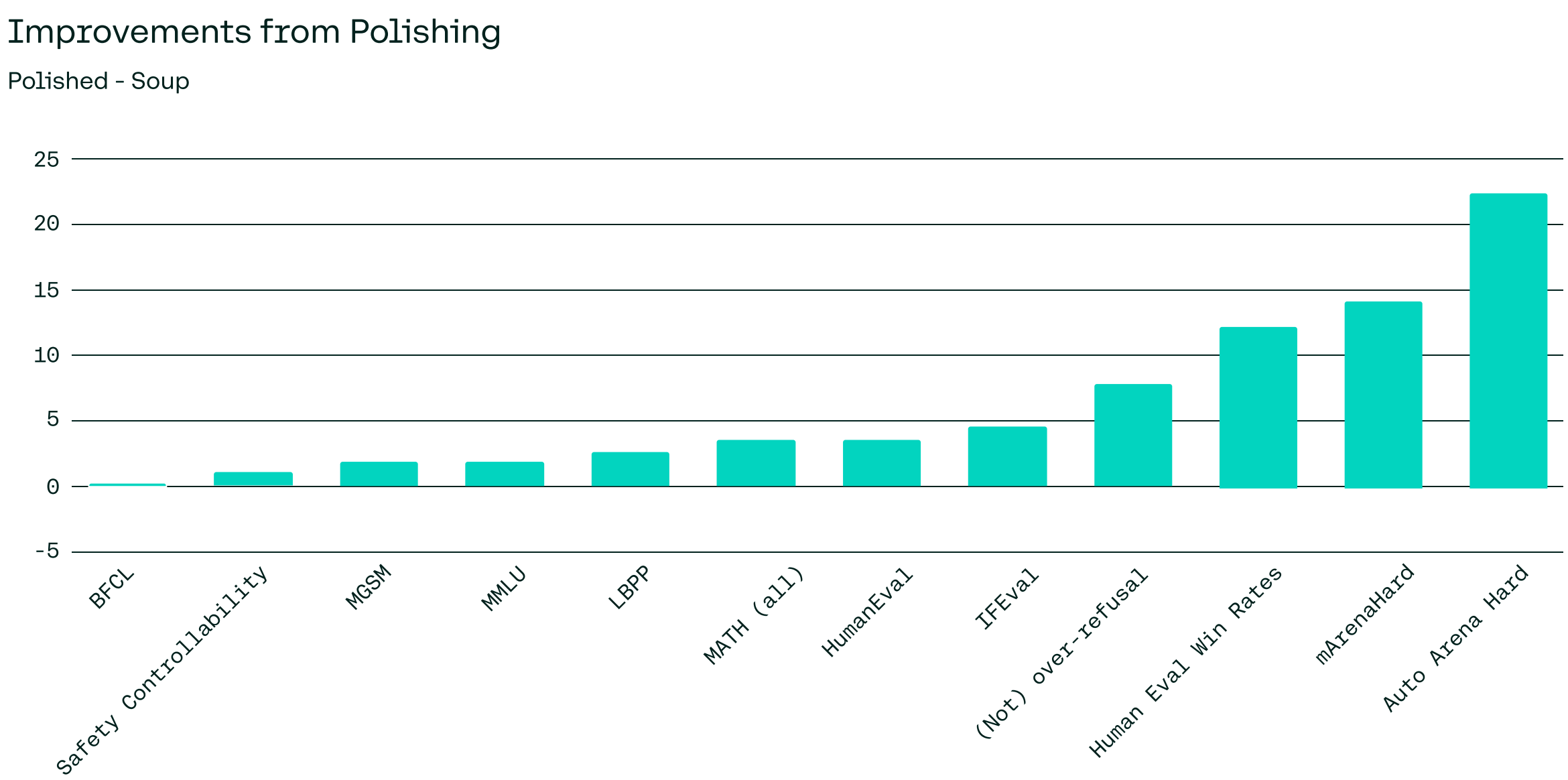

Figure 14: Performance improvements across various domains during polishing between the inital RL soup model and the polished model.

## 4.11 Human Evaluation

While automated evaluation benchmarks provide quick feedback and allow for efficient hill-climbing in specific task settings, automatically assessing the perceived quality of models remains challenging (Ni et al., 2024). To validate broader model performance, our most promising model candidates are additionally evaluated by human annotators. This section describes our human evaluation setup, including details on the curation of our internal evaluation dataset, an overview of our annotation strategy, as well as the results for Command A in head-to-head evaluation against competitors.

#### 4.11.1 Evaluation Data

Chatbot Arena (Chiang et al., 2024) is a popular LLM benchmark that involves crowd-sourced human annotations. The framework relies on human preference judgments between two model-generated completions over user-provided prompts, from which the authors then derive Elo scores to build up-to-date leaderboards (Boubdir et al., 2024). While Chatbot Arena-like evaluation provides an extremely useful quality signal on *user-perceived model quality*, it has several drawbacks for the efficient evaluation of internal model candidates. Two challenges are the long ramp-up time required to provide performance signal on a new model (often requiring multiple thousands of ratings before producing a reliable Elo score), as well as the reliance on user-provided prompts that often end up skewing towards simpler topics. In order to provide a more immediate and targeted feedback signal on human-perceived model performance, we instead curate a static collection of single-turn prompts for the internal evaluation of our models versus competitors.

Prompts in our internal collection are primarily curated from scratch by our pool of annotators to avoid accidental contamination for competitor models by re-using existing prompt collections. Instructions are specifically targeted towards creating complex real-world prompts that span a broad range of typical LLM-assisted tasks. For our purposes, we define complexity as a function of the number of different asks contained in a single prompt. As an example, asking for a summary with a certain length requirement would be assigned a complexity score of 2. The first point for invoking a particular task (summarisation), and the second point for additionally restricting the output (length constraint). We automatically filter prompts with the help of Command R+ Refresh along several axes, building up a more comprehensive notion of prompt complexity. While the filtering may be overly strict in some instances, our findings show that the resulting pool of prompts is sufficiently difficult even for state-of-the-art models. We prune the final dataset to around 800 prompts, further subset into "general" (350 prompts), "reasoning" (150 prompts), and "code" (300 prompts).

The "general" split focuses on general-purpose tasks that do not require deep technical expertise. This can include open-ended document generation or idea generation requests, and at most requires basic understanding of formatting languages such as CSV or JSON. The "reasoning" split includes more reasoning- and math-heavy prompts that generally require undergraduate-level understanding of one STEM discipline. For the "code" split, we specifically instruct annotators to curate prompts across a number of target programming languages, with a focus on analysis and debugging rather than code generation.

#### 4.11.2 Annotation Methodology

Our annotation process comprises a relatively straightforward pairwise preference evaluation setup. For each annotation task, we present annotators with one prompt and two completions from competing models. Annotators are first tasked with evaluating the quality of each completion *separately*, assigning a score from 1 (flawed) to 5 (perfect). They are then given the option to label common failure modes. Finally, annotators provide their preference between the two completions. The choices correspond to "Completion A is much better", "Completion A is slightly better", "Tie", "Completion B is slightly better", and "Completion B is much better". To avoid positional bias, we randomly shuffle the order in which completions are shown to annotators. We compare different models' performance based on their win rate versus a fixed competitor model. We assign the win rate in a pairwise matchup as a single score, distributing ties, as:

$$winrate = \frac{wins \ + \ (0.5 \ \times ties)}{wins + ties + losses}$$

We group the strong and weak preferences for wins or losses together and find that computing a win rate over the 5-point scale does not change model rankings, but helps annotators gain confidence in their given preference ratings. For a complete pairwise evaluation of 800 samples across all three subsets, on average 65 annotators contribute to a single evaluation run.

#### 4.11.3 Results

Table 23 shows the results for pairwise human annotation runs against different competitor models. [11] Our results show that Command A is competitive with frontier models on both general and reasoning subsets.

[11]We obtain completions for GPT-family models directly through the OpenAI API. We use the TogetherAI endpoints for Llama 3.3 and DeepSeek V3.

| vs. | Command A Win Rate (%) General | Reasoning | Code |
|---|---|---|---|
| GPT-4o | 50.4 | 51.4 | 46.8 |
| GPT-4.5 Preview | 47.2 | 30.7 | 38.3 |
| DeepSeek V3 | 49.0 | 49.3 | 54.7 |
| Llama 3.3 70B Instruct | 68.8 | 71.7 | 63.4 |
| Llama 3.1 405B Instruct | 61.6 | 64.0 | 61.6 |

Table 23: Win rate of human-annotated pairwise evaluation between Command A and different competitor models on our internal test set. Win rates are from the perspective of Command A (50% is tie, higher numbers mean Command A wins by a larger margin).

Interestingly, we also observe that GPT-4.5-preview improves significantly over its predecessor on reasoning-heavy prompts.

Annotators strongly prefer Command A over Llama 3.3 70B Instruct for all subsets, including code. Our analysis across evaluation results indicates that humans particularly prefer Command A's ability to respect requests for a particular formatting or style.

To further illustrate the importance of polishing as a critical step towards human preference optimisation (see Section 3.5), we compare the human evaluation results of an earlier internal checkpoint, a result of the expert merging before polishing. Evaluating both our earlier checkpoint and the final model candidate against GPT-4o, we see substantial gains as reflected in human preference. For our final model, we manage an absolute improvement of more than seven percentage points in win rate on the general subset (43.2 $\rightarrow$ 50.4), 10 points on reasoning (41.4 $\rightarrow$ 51.4), and almost 17 points (30.0 $\rightarrow$ 46.8) on code.

Overall, Command A is significantly more preferred by human evaluators to models such as Llama 3.1 405B Instruct across all subsets, and is competitive with state-of-the-art models such as GPT-4o and DeepSeek V3 while being considerably more efficient.

# 5 Conclusion

This technical report detailed the development of Command A, shared extensive performance evaluations across many domains and languages, and shared additional results for Command R7B. Command A represents a significant advancement in LLMs for enterprise, achieving best-in-class performance across a wide range of tasks with optimal efficiency. Our models excel in enterprise-relevant tasks such as agentic workflows, multilingual understanding and generation, and instruction-following. Key innovations introduced include data and architectural optimisations, self-refinement algorithms, and a model merging-based approach that ensures expert-level performance across diverse capabilities within a single model.

Command A outperforms comparable models in both efficiency and computational overhead, requiring fewer resources for serving, making it easy to deploy on-premises or in private cloud environments on just two A100 or H100 GPUs, and delivering tokens at a higher rate. The release of model weights under a non-commercial license further facilitates community-based exploration and research. Command A sets a new standard for LLMs in enterprise applications, balancing performance, efficiency, and versatility — and providing maximum performance for minimal compute.

# A Authors

Team Cohere: Aakanksha, Arash Ahmadian, Marwan Ahmed, Jay Alammar, Milad Alizadeh, Yazeed Alnumay, Sophia Althammer, Arkady Arkhangorodsky, Viraat Aryabumi, Dennis Aumiller, Raphaël Avalos, Zahara Aviv, Sammie Bae, Saurabh Baji, Alexandre Barbet, Max Bartolo, Björn Bebensee, Neeral Beladia, Walter Beller-Morales, Alexandre Bérard, Andrew Berneshawi, Anna Bialas, Phil Blunsom, Matt Bobkin, Adi Bongale, Sam Braun, Maxime Brunet, Samuel Cahyawijaya, David Cairuz, Jon Ander Campos, Cassie Cao, Kris Cao, Roman Castagné, Julián Cendrero, Leila Chan Currie, Yash Chandak, Diane Chang, Giannis Chatziveroglou, Hongyu Chen, Claire Cheng, Alexis Chevalier, Justin T. Chiu, Eugene Cho, Eugene Choi, Eujeong Choi, Tim Chung, Volkan Cirik, Ana Cismaru, Pierre Clavier, Henry Conklin, Lucas Crawhall-Stein, Devon Crouse, Andres Felipe Cruz-Salinas, Ben Cyrus, Daniel D'souza, Hugo Dalla-Torre, John Dang, William Darling, Omar Darwiche Domingues, Saurabh Dash, Antoine Debugne, Théo Dehaze, Shaan Desai, Joan Devassy, Rishit Dholakia, Kyle Duffy, Ali Edalati, Ace Eldeib, Abdullah Elkady, Sarah Elsharkawy, Irem Ergün, Beyza Ermis, Marzieh Fadaee, Boyu Fan, Lucas Fayoux, Yannis Flet-Berliac, Nick Frosst, Matthias Gallé, Wojciech Galuba, Utsav Garg, Matthieu Geist, Mohammad Gheshlaghi Azar, Ellen Gilsenan-McMahon, Seraphina Goldfarb-Tarrant, Tomas Goldsack, Aidan Gomez, Victor Machado Gonzaga, Nithya Govindarajan, Manoj Govindassamy, Nathan Grinsztajn, Nikolas Gritsch, Patrick Gu, Shangmin Guo, Kilian Haefeli, Rod Hajjar, Tim Hawes, Jingyi He, Sebastian Hofstätter, Sungjin Hong, Sara Hooker, Tom Hosking, Stephanie Howe, Eric Hu, Renjie Huang, Hemant Jain, Ritika Jain, Nick Jakobi, Madeline Jenkins, JJ Jordan, Dhruti Joshi, Jason Jung, Trushant Kalyanpur, Siddhartha Rao Kamalakara, Julia Kedrzycki, Gokce Keskin, Edward Kim, Joon Kim, Wei-Yin Ko, Tom Kocmi, Michael Kozakov, Wojciech Kryściński, Arnav Kumar Jain, Komal Kumar Teru, Sander Land, Michael Lasby, Olivia Lasche, Justin Lee, Patrick Lewis, Jeffrey Li, Jonathan Li, Hangyu Lin, Acyr Locatelli, Kevin Luong, Raymond Ma, Lukáš Mach, Marina Machado, Joanne Magbitang, Brenda Malacara Lopez, Aryan Mann, Kelly Marchisio, Olivia Markham, Alexandre Matton, Alex McKinney, Dominic McLoughlin, Jozef Mokry, Adrien Morisot, Autumn Moulder, Harry Moynehan, Maximilian Mozes, Vivek Muppalla, Lidiya Murakhovska, Hemangani Nagarajan, Alekhya Nandula, Hisham Nasir, Shauna Nehra, Josh Netto-Rosen, Daniel Ohashi, James Owers-Bardsley, Jason Ozuzu, Dennis Padilla, Gloria Park, Sam Passaglia, Jeremy Pekmez, Laura Penstone, Aleksandra Piktus, Case Ploeg, Andrew Poulton, Youran Qi, Shubha Raghvendra, Miguel Ramos, Ekagra Ranjan, Pierre Richemond, Cécile Robert-Michon, Aurélien Rodriguez, Sudip Roy, Sebastian Ruder, Laura Ruis, Louise Rust, Anubhav Sachan, Alejandro Salamanca, Kailash Karthik Saravanakumar, Isha Satyakam, Alice Schoenauer Sebag, Priyanka Sen, Sholeh Sepehri, Preethi Seshadri, Ye Shen, Tom Sherborne, Sylvie Shang Shi, Sanal Shivaprasad, Vladyslav Shmyhlo, Anirudh Shrinivason, Inna Shteinbuk, Amir Shukayev, Mathieu Simard, Ella Snyder, Ava Spataru, Victoria Spooner, Trisha Starostina, Florian Strub, Yixuan Su, Jimin Sun, Dwarak Talupuru, Eugene Tarassov, Elena Tommasone, Jennifer Tracey, Billy Trend, Evren Tumer, Ahmet Üstün, Bharat Venkitesh, David Venuto, Pat Verga, Maxime Voisin, Alex Wang, Donglu Wang, Shijian Wang, Edmond Wen, Naomi White, Jesse Willman, Marysia Winkels, Chen Xia, Jessica Xie, Minjie Xu, Bowen Yang, Tan Yi-Chern, Ivan Zhang, Zhenyu Zhao, and Zhoujie Zhao.

## A.1 Acknowledgements

We would also like to acknowledge the contributions of the following people who helped make this work possible: Robert Li, Olivier Pietquin, Karina Waluk, Bill Wu, and James Zhou. We would also like to thank our talented team of internal annotators and model trainers.

This technical report was written with the assistance of the models described in this technical report.

# B Appendices

## B.1 Instruction-Following

We use the following custom preamble override for internal IFEval evaluation for Command A:

> *You are in non-interactive mode. Please be as comprehensive and accurate as possible and do not introduce your response and / or ask follow-up questions.*

## B.2 Multilingual

### B.2.1 Expert Training Considerations

**Supervised Fine-Tuning (SFT).** We employ all SFT datasets as mentioned in Section 3.3.3.1. To train the model, we use the Adam optimiser with a peak learning rate of $2.5 \times 10^{-5}$, cosine decay to $1.25 \times 10^{-5}$, $\beta_1 = 0.9$, $\beta_2 = 0.95$ and a weight decay of 0.1. We merge several models with the same configuration (but a different random seed) at this stage.

**Preference Tuning.** Following the SFT stage, we perform offline preference tuning using both human-annotated and synthetically generated multilingual preference data (with best-of-N or machine translation). We use DPO with the Adam optimiser, a peak learning rate of $1.25 \times 10^{-5}$ decaying to $1.25 \times 10^{-6}$, $\beta = 0.2$ and use SFT regularisation with the same data mixture we used in the SFT stage. We do not merge over multiple seeds at this stage.

### B.2.2 Results

We show additional results on MGSM in Table 24, and INCLUDE-44 in Table 25 — highlighting the highly competitive multilingual capabilities of our models.

| | Avg. | de | es | fr | ja | ru | zh |
|---|---|---|---|---|---|---|---|
| Command A | 90.1 | 92.0 | 92.4 | 85.1⋆ | 87.1 | 94.0 | 90.0 |
| DeepSeek V3 | **92.3** | 92.4 | 96.0 | **90.4** | 89.6 | 94.0 | 91.6 |
| Claude 3.7 Sonnet | 92.1 | **93.6** | **96.4** | 85.5 | 89.2 | **95.6** | 92.4 |
| Llama 3.3 70B Instruct | 91.5 | 92.8 | 91.6 | 90.0 | 89.6 | 95.2 | 90.0 |
| Qwen 2.5 72B Instruct Turbo | 90.5 | 92.0 | 94.4 | 87.6 | 87.1 | 92.4 | 89.6 |
| Mistral Large 2 | 90.1 | 90.0 | 90.4 | 85.9 | **90.3** | 92.8 | 91.5 |
| Gemini 2.0 Flash | 90.0 | 90.8 | 92.8 | 83.1 | 87.1 | 95.2 | 90.8 |
| GPT-4o | 89.6 | 92.0 | 90.8 | 83.1 | 87.1 | 92.0 | **92.8** |
| Gemini 1.5 Pro | 88.4 | 91.2 | 93.2 | 77.5 | 86.7 | 92.4 | 89.6 |
| Llama 3.1 405B Instruct | 74.1 | 89.8 | 82.4 | 7.7 | 85.8 | 90.3 | 88.6 |
| Command R7B | 75.0 | 77.1 | 78.3 | 71.5 | 66.7 | 79.9 | 76.3 |
| Gemma 2 9B Instruct | **80.7** | **83.1** | **85.9** | **76.7** | 72.3 | **86.3** | 79.9 |
| Gemini 1.5 Flash-8B | 80.3 | 79.9 | 80.7 | **76.7** | **79.1** | 84.3 | 80.7 |
| Claude 3 Haiku | 76.9 | 77.9 | 77.9 | 73.5 | 74.7 | 80.3 | 77.1 |
| Ministral 8B | 76.1 | 81.0 | 80.6 | 74.6 | 64.4 | 81.1 | 74.7 |
| Llama 3.1 8B Instruct | 70.4 | 73.9 | 67.1 | 65.9 | 62.7 | 77.9 | 75.1 |
| Qwen 2.5 7B Instruct Turbo | 70.3 | 81.9 | 75.9 | 27.3 | 72.7 | 82.3 | **81.5** |

Table 24: MGSM scores using simple-evals. ⋆: we did not train with the simple-evals template and our French outputs sometimes contain a comma (used as a decimal separator), which simple-evals counts as wrong. With our internal implementation of MGSM, we score 90.0% on French.

| | Avg. | ar | de | el | es | fa | fr | he | hi | id | it | ja | ko | nl | pl | pt | ru | tr | uk | vi | zh |
|---|---|---|---|---|---|---|---|---|---|---|---|---|---|---|---|---|---|---|---|---|---|
| Command A | 74.3 | 73.2 | 67.6 | 68.0 | 79.3 | 61.3 | 77.6 | 75.5 | 67.6 | 72.7 | 86.5 | 86.8 | 72.4 | 81.3 | 78.6 | 77.5 | 69.2 | 67.7 | 76.5 | 70.7 | 75.8 |
| Claude 3.7 Sonnet | **81.0** | 79.2 | 66.9 | **74.7** | **85.1** | **73.4** | **84.2** | **87.8** | 77.3 | **81.6** | **91.4** | 93.4 | 73.6 | **87.1** | **87.0** | **80.9** | **78.8** | 68.4 | **87.1** | **80.5** | 80.7 |
| Gemini 2.0 Flash | 78.8 | 79.0 | 69.1 | 71.8 | 83.1 | 64.6 | 79.5 | 83.3 | **80.4** | 77.8 | 89.8 | 91.2 | **79.4** | 86.0 | 85.0 | 79.1 | 73.7 | 66.2 | 80.4 | 77.5 | 78.3 |
| GPT-4o | 78.7 | **79.3** | 68.3 | 72.5 | 83.1 | 63.3 | 82.3 | 83.1 | 76.4 | 78.2 | 89.2 | **93.6** | 73.8 | 85.8 | 82.7 | 78.9 | 74.8 | 66.2 | 82.5 | 79.5 | 79.3 |
| DeepSeek V3 | 76.9 | 73.9 | 64.7 | 66.5 | 82.5 | 63.5 | 83.1 | 80.2 | 75.0 | 76.5 | 88.5 | 91.8 | 71.4 | 83.7 | 80.3 | 77.9 | 72.8 | 66.8 | 77.8 | 78.0 | 83.9 |
| Gemini 1.5 Pro | 76.4 | 72.5 | 66.2 | 70.7 | 79.6 | 65.3 | 79.4 | 78.2 | 77.1 | 76.0 | 87.2 | 90.8 | 72.3 | 86.2 | 80.4 | 75.7 | 71.7 | 67.7 | 78.5 | 75.1 | 77.0 |
| Llama 3.1 405B Instruct | 75.3 | 70.7 | **71.7** | 62.2 | 81.4 | 60.0 | 77.4 | 77.8 | 73.9 | 76.4 | 76.8 | 88.5 | 68.4 | 85.8 | 84.7 | 75.8 | 72.0 | 71.9 | 76.5 | 77.6 | 76.9 |
| Llama 3.3 70B Instruct | 75.0 | 71.7 | 65.5 | 64.7 | 78.0 | 59.5 | 75.2 | 78.5 | 72.4 | 76.4 | 89.2 | 85.2 | 67.0 | 83.5 | 86.5 | 77.9 | 69.6 | **73.0** | 74.5 | 76.4 | 74.9 |
| Qwen 2.5 72B Instruct Turbo | 72.6 | 70.5 | 64.0 | 57.1 | 76.9 | 57.1 | 77.3 | 70.5 | 71.5 | 73.1 | 85.8 | 86.0 | 70.6 | 84.0 | 71.9 | 76.2 | 69.4 | 64.8 | 71.3 | 70.0 | **84.4** |
| Mistral Large 2 | 72.0 | 68.3 | 66.2 | 66.0 | 79.5 | 55.3 | 76.6 | 74.9 | 69.3 | 72.2 | 87.0 | 85.8 | 65.0 | 82.4 | 74.3 | 75.3 | 68.3 | 65.3 | 71.8 | 66.2 | 71.0 |
| Command R7B | 55.8 | 58.0 | 51.8 | 41.3 | 64.0 | 41.4 | 58.7 | 53.6 | 47.2 | 58.4 | 69.2 | 61.9 | 58.2 | 59.5 | **71.0** | 49.9 | 47.3 | 62.0 | 58.9 | 52.2 | 51.9 |
| Gemini 1.5 Flash-8B | **66.0** | 61.4 | 54.7 | **61.7** | 71.2 | **49.3** | **69.4** | **68.9** | **65.4** | **67.8** | **79.4** | **77.4** | **64.0** | **74.6** | 67.5 | **67.3** | 62.1 | **63.0** | 66.5 | **63.1** | 65.9 |
| Claude 3 Haiku | 64.0 | **65.9** | 50.4 | 54.2 | **71.5** | 46.9 | 67.1 | 68.0 | 60.1 | 67.1 | 76.6 | 75.0 | 62.8 | 72.6 | 69.3 | 64.1 | **62.3** | 58.6 | **69.3** | 56.0 | 61.3 |
| Gemma 2 9B Instruct | 61.7 | 57.1 | **58.3** | 55.3 | 65.5 | 44.9 | 64.0 | 66.5 | 58.1 | 63.6 | 72.4 | 73.7 | 56.2 | 70.2 | 64.1 | 65.0 | 57.8 | 59.3 | 66.0 | 60.2 | 56.5 |
| Qwen 2.5 7B Instruct Turbo | 59.9 | 57.1 | 53.2 | 49.3 | 65.8 | 43.4 | 63.5 | 58.4 | 56.7 | 63.5 | 72.8 | 71.9 | 61.4 | 69.7 | 59.5 | 61.5 | 52.5 | 52.6 | 57.1 | 56.4 | **72.8** |
| Llama 3.1 8B Instruct | 54.9 | 55.1 | 53.2 | 33.5 | 62.7 | 43.8 | 59.2 | 54.5 | 49.2 | 60.5 | 70.6 | 58.3 | 50.2 | 63.7 | 55.1 | 59.5 | 49.1 | 56.0 | 54.7 | 55.5 | 53.6 |
| Ministral 8B | 50.2 | 43.8 | 42.4 | 41.6 | 59.5 | 38.7 | 57.0 | 49.5 | 46.6 | 49.5 | 57.5 | 57.3 | 51.0 | 58.4 | 55.8 | 49.2 | 51.1 | 46.7 | 54.0 | 44.5 | 50.6 |

Table 25: INCLUDE-44 scores for individual languages.

## B.3 Code

### B.3.1 Expert Training Considerations

**Stage 1 large-scale supervised learning.** This expert is trained using Adam optimisation, a learning rate peak of $5{\times}10^{-5}$, cosine decay to $5{\times}10^{-6}$, beta values ($\beta_1 = 0.9, \beta_2 = 0.95$), weight decay factor of 0.1, and gradient norm clipping at peak 1.0. Our regularisation is inspired by similar code expert training such as Qwen 2.5 Coder (Hui et al., 2024).

**Stage 2 supervised learning on only high-quality data.** We follow a similar schedule for fine-tuning of the merged model described in Stage 1.

**Stage 3 RL over scored or preference-pair data.** We use a constant learning rate of $2 \times 10^{-6}$, a regularisation parameter $\beta$ of 0.06, and otherwise match the hyperparameters used for SFT.

### B.3.2 Results

We additionally show a breakdown of results across programming languages for HumanEval and Less Basic Python Problems (LBPP) in Table 26.

| | **HumanEval** | | | | | | | | **Less Basic Python Problems (LBPP)** | | | | | | |
|---|---|---|---|---|---|---|---|---|---|---|---|---|---|---|---|
| | Avg. | Python | C++ | Rust | Java | JavaScript | Go | COBOL | Avg. | Python | C++ | Rust | Java | JavaScript | Go |
| Command A | 76.2 | 85.4 | 81.1 | 73.2 | 86.0 | 64.6 | 72.0 | 25.3 | 51.5 | 58.4 | 50.3 | 41.6 | 50.0 | 58.2 | 50.3 |
| Command A Expert | 77.5 | 86.6 | 76.8 | **74.4** | 76.2 | 79.9 | 71.3 | **29.8** | 50.8 | 59.6 | 51.6 | 39.6 | 53.2 | 54.3 | 46.6 |
| Command A Agentic | — | — | — | — | — | — | — | — | **65.4** | **71.4** | **68.8** | **57.7** | 63.9 | **68.0** | **62.7** |
| Command R7B | 50.7 | 61.0 | 54.9 | 39.0 | 52.4 | 53.7 | 43.3 | 7.0 | 21.9 | 26.1 | 21.1 | 13.4 | 23.4 | 29.4 | 17.4 |
| Command R Refresh | 54.7 | 67.1 | 56.1 | 47.6 | 59.2 | 64.6 | 33.5 | 1.9 | 24.7 | 33.5 | 25.5 | 16.1 | 22.2 | 26.8 | 23.6 |
| Command R+ Refresh | 54.4 | 72.0 | 54.3 | 50.6 | 67.1 | 62.8 | 19.5 | 2.5 | 25.6 | 38.5 | 26.7 | 21.5 | 30.4 | 28.1 | 8.1 |
| Llama 3.3 70B Instruct | 75.5 | 80.5 / 85.4 | 75.0 | 63.4 | 83.4 | 81.7 | 62.2 | 3.2 | 47.8 | 55.3 | 42.2 | 32.9 | 56.3 | 56.2 | 43.5 |
| Mistral Large 2 | 82.9 | **94.5** | 86.0 | 72.6 | 84.8 | **87.2** | 72.6 | 10.8 | 54.0 | 62.7 | 53.4 | 36.2 | 51.3 | 52.9 | 42.2 |
| Qwen 2.5 72B Instruct | 78.5 | 86.6 / 86.6 | 76.8 | 72.0 | 82.9 | 81.1 | 71.3 | 6.3 | 48.3 | 59.6 | 46.6 | 32.9 | 49.4 | 60.8 | 39.8 |
| Llama 3.1 405B Instruct | 76.7 | 89.0 / 88.4 | 81.7 | 53.7 | **87.8** | 79.3 | 69.5 | 3.2 | 52.7 | 59.6 | 50.9 | 38.9 | 57.6 | 61.4 | 47.2 |
| DeepSeek V3 | **83.5** | 92.1 | **90.2** | 70.7 | 85.4 | 85.4 | **77.4** | 15.2 | 61.5 | 67.1 | 59.0 | **57.1** | **67.1** | 66.0 | 52.8 |

Table 26: Full `pass@1` results for HumanEval and LBPP across Python, C++, Rust Java, Javascript, Go. and COBOL. All results are internal reproductions using an identical prompt except where '/' indicates an external value first, and footnote citation, and the internal reproduction second. Average (Avg.) results are a sample-weighted average across Python, C++, Rust Java, Javascript and Go languages.

Sources of externally reported numbers in all code-related tables in this report are as follows:

**Code Understanding Benchmarks** (Table 12)

- Llama 3.3 70B Instruct: https://github.com/meta-llama/llama-models/blob/main/models/lla

ma3_3/MODEL_CARD.md, https://bigcode-bench.github.io/
- Mistral Large 2: https://livecodebench.github.io/
- Qwen 2.5 72B Instruct: https://bigcode-bench.github.io/
- Llama 3.1 405B Instruct: https://github.com/meta-llama/llama-models/blob/main/models/llama3_3/MODEL_CARD.md
- DeepSeek V3: https://livecodebench.github.io/, https://bigcode-bench.github.io/

**Code Editing Benchmarks** (Table 13)
- DeepSeek V3: https://www.deepseek.com/

**HumanEval and LBPP** (Table 26)
- Llama 3.3 70B Instruct and Llama 3.1 405B Instruct: https://github.com/meta-llama/llama-models/blob/main/models/llama3_3/MODEL_CARD.md
- Qwen 2.5 72B Instruct: From Hui et al. (2024).

## B.4 Reasoning

**Supervised Fine-Tuning.** We train using the Adam optimiser with peak learning rate of $2.5\times10^{-5}$, cosine decay to $2.5 \times 10^{-6}$, $\beta_1 = 0.9$, $\beta_2 = 0.95$, weight decay of 0.01, and gradient norm clipping peak at 1.0.

**Preference Tuning.** We train using CoPG with the Adam optimiser, using a learning rate of $2\times10^{-6}$ with no decay, $\beta_1 = 0.9$, $\beta_2 = 0.95$, and gradient norm clipping peak at 1.0.

## B.5 Long Context

Training is conducted using the Adam optimiser with a peak learning rate of $2.5 \times 10^{-5}$, cosine decay to $2.5 \times 10^{-6}$, $\beta_1 = 0.9$, $\beta_2 = 0.95$, weight decay of 0.01, and gradient norm clipping peak at 1.0.

## B.6 Safety

| | Contextual Over-Refusal | Contextual Accuracy | Strict Over-Refusal | Strict Accuracy |
|---|---|---|---|---|
| Command A | 8.3 | 75.4 | 10.2 | 87.5 |
| Claude 3.5 Sonnet | 10.1 | **82.5** | 10.4 | **89.6** |
| DeepSeek V3 | **3.6** | 66.6 | 1.7 | 68.7 |
| GPT-4o | 21.3 | 77.8 | 10.4 | 86.9 |
| Llama 3.1 405B | 5.9 | 70.3 | 5.8 | 80.5 |
| Llama 3.3 70B | **3.6** | 60.9 | 4.0 | 67.9 |
| Mistral Large Latest | 8.9 | 70.6 | 8.7 | 81.1 |
| Qwen 2.5 72B | 10.1 | 72.2 | **1.2** | 78.3 |
| Command R+ Refresh | 8.9 | 61.8 | 12.7 | 70.0 |
| Command R7B | 14.2 | 60.9 | 7.5 | 62.3 |
| Gemini 2.0 Flash | 26.6 | 69.7 | 14.7 | 79.0 |
| Gemma 2 9B | 39.1 | **72.5** | 14.5 | **81.8** |
| Llama 3.1 8B | **4.1** | 70.7 | **6.9** | 81.0 |
| Qwen 2.5 7B | 18.3 | 70.3 | 8.7 | 74.7 |
| Command R Refresh | 9.5 | 63.1 | 10.4 | 68.0 |

Table 27: Safety mode performance compared to similarly sized models. We pass the instructions for each safety mode to competitors as system message or first message to allow for the mode comparison. Note that the over-refusal set was developed by red-teaming Command R+ Refresh, and therefore is specifically challenging for Command A models. The top performing model for each size category is bolded in each column. Higher accuracy and lower over-refusal rates correspond to better performance.

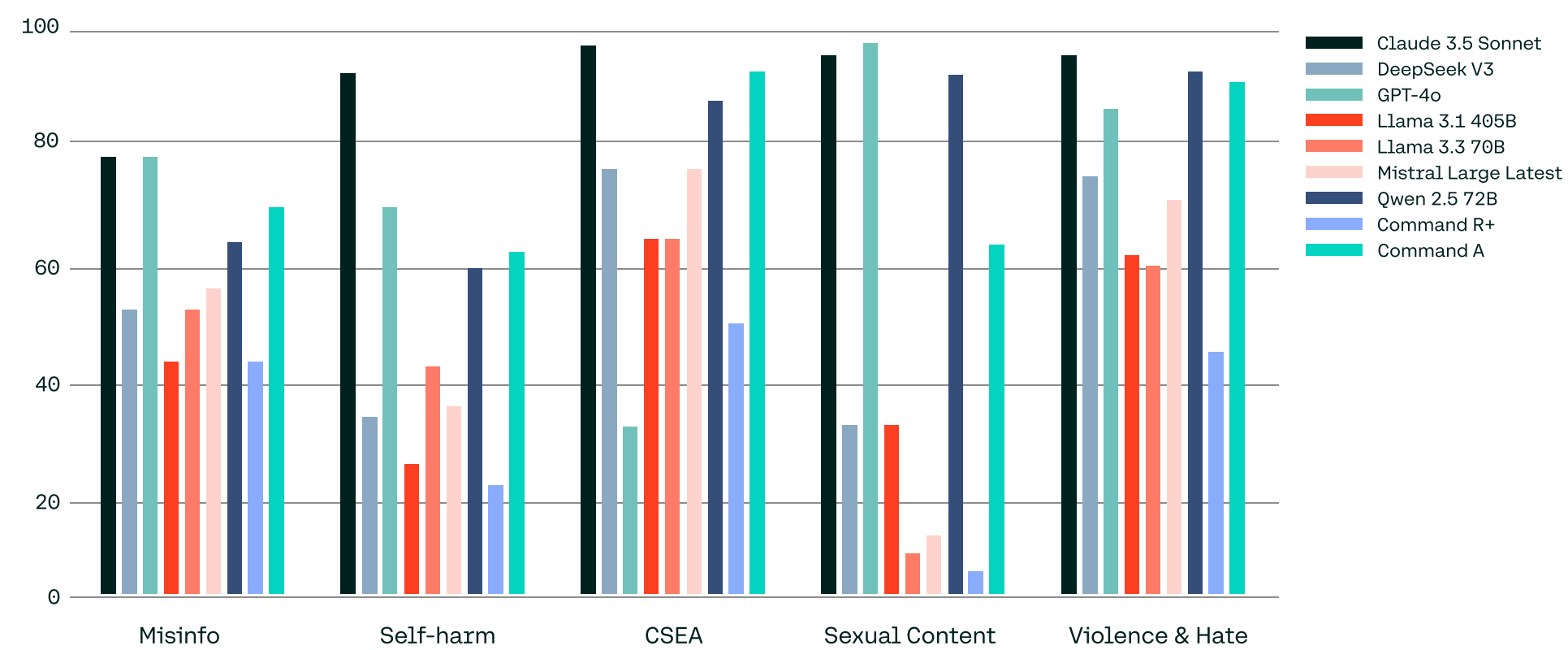

Figure 15: Absolute safety performance for large models in the default setting.

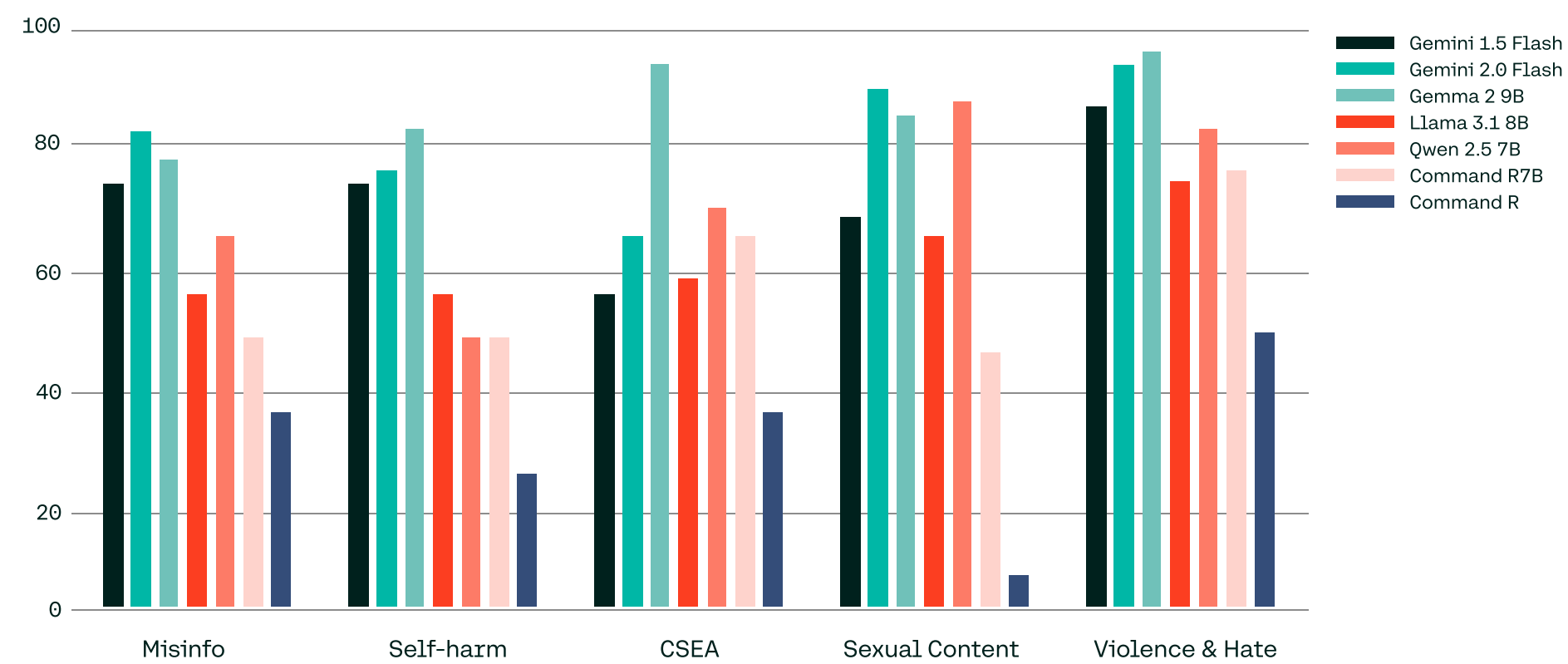

Figure 16: Absolute safety performance for small models in the default setting.

### B.6.1 Expert Training Considerations

**Supervised fine-tuning.** During the SFT stage, we train the safety expert using the Adam optimiser with $\beta_1 = 0.9$, $\beta_2 = 0.95$, and a peak learning rate of $10^{-4}$ decayed to $10^{-5}$ with a cosine schedule. Gradient norm is clipped to 1, and weight decay is weighted by $10^{-3}$.

**Offline preference tuning.** We train the Safety expert using the same hyper-parameters as above, except the peak learning rate is $10^{-6}$ decayed to $10^{-7}$. We use IPO with a KL regularisation parameter, $\beta = 0.03$.

### B.6.2 Results

Table 27 provides additional results for the safety mode performance of the Command A models, while Figures 15 and 16 show absolute safety performance for large and small models respectively in the default safety setting, further highlighting our models' competitive safety performance.

## B.7 Evaluation on Standard Benchmarks

We provide further details about how we measure performance on standard benchmarks:

- **MMLU** (Hendrycks et al., 2021) measures university-level academic knowledge across a diverse range of subjects. We run 5-shot, with Chain-of-Thought (CoT).
- **MMLU-Pro** (Wang et al., 2024b) is an enhanced version of MMLU, designed to evaluate knowledge across a diverse range of professional and academic domains, including law, medicine, and engineering.

We run 5-shot, with CoT.

- **GPQA** (Rein et al., 2023) measures graduate-level academic knowledge in specialized STEM topics. We run on 0-shot, with CoT, and we report results only on the diamond subset.
- **IFEval** (Zhou et al., 2023) measures instruction-following ability across 25 types of verifiable instructions (e.g. output length, keyword inclusion/exclusion, formatting). We compute the average of the prompt-level strict accuracy (i.e., the fraction of dataset prompts where all verifiable instructions are followed) and the instruction-level strict accuracy (i.e., the fraction of verifiable instructions that are followed, this allows partial credit as most prompts include multiple instructions).
- **InFoBench** (Qin et al., 2024) also measures instruction-following across five broad types of instructions: content, linguistic, style, format, and number. Each prompt in InFoBench is paired with a set of yes-no evaluation questions, and we use GPT-4o to answer these questions. We compute the overall average accuracy across the fraction of correctly answered questions per prompt.